%% file: main.tex
\documentclass[10pt,twocolumn,letterpaper]{article}

\usepackage[pagenumbers]{wacv} % To force page numbers, e.g. for an arXiv version
\usepackage{times}
\usepackage{epsfig}
\usepackage{graphicx}
\usepackage{amsmath}
\usepackage{amssymb}
\usepackage[accsupp]{axessibility}

\input{packages.tex}

\usepackage[pagebackref, colorlinks, breaklinks=true, bookmarks=false]{hyperref}
\usepackage[capitalize]{cleveref}

\hypersetup{
     pdftitle={Differentiable JPEG: The Devil is in the Details},
     pdfsubject={Differentiable Image Compression},
     pdfauthor={Christoph Reich, Biplob Debnath, Deep Patel, and Srimat Chakradhar}
}

\newcommand{\q}{\mathrm{q}}
\newcommand{\qt}{\mathbf{QT}}
\newcommand{\jpeg}{\operatorname{JPEG}}

\DeclareMathOperator{\sign}{sign}

\newcommand \colorindicator[2]{%
	#1~{\textcolor{#2}{$\blacksquare\!\!\!\!\!\blacksquare$}}%
}

\renewcommand*{\paragraph}[1]{\smallskip\noindent\textbf{#1}\hspace{0.25em}}

\def\vs{\emph{vs}\onedot} 

\DeclareSIUnit \flop {FLOP}

\usepackage[capitalize]{cleveref}
\crefname{section}{Sec.}{Secs.}
\Crefname{section}{Section}{Sections}
\Crefname{table}{Table}{Tables}
\crefname{table}{Tab.}{Tabs.}

\input{colors.tex}

\DeclareSIUnit\fps{fps}

\def\aspace{\hspace{1.25cm}}

\def\wacvPaperID{708} % *** Enter the WACV Paper ID here
\def\confName{WACV}
\def\confYear{2024}

\usepackage{fancyhdr}
\usepackage{setspace}

\fancypagestyle{firststyle}
{

	\fancyhf{}
	\lfoot{{\footnotesize\begin{spacing}{.5}\parbox{\linewidth}{\vspace{2em}
					To appear in Proceedings of the \emph{IEEE/CVF Winter Conference on Applications of Computer Vision (WACV)}, 2024.%
					\\\hrule\vspace{\baselineskip}
					\copyright~2023 IEEE. Personal use of this material is permitted. Permission from IEEE must be obtained for all other uses, in any current or future media, including reprinting/republishing this material for advertising or promotional purposes, creating new collective works, for resale or redistribution to servers or lists, or reuse of any copyrighted component of this work in other works. 
	}\end{spacing}}}
}

\begin{document}
	
	%%%%%%%%% TITLE
	\title{Differentiable JPEG: The Devil is in the Details}
	
	\author{Christoph Reich\textsuperscript{1,2}
		\aspace Biplob Debnath\textsuperscript{1}
		\aspace Deep Patel\textsuperscript{1}
		\aspace Srimat Chakradhar\textsuperscript{1}\\
		\textsuperscript{1}\hspace{0.02cm}NEC Laboratories America, Inc. \hspace{0.75cm} \textsuperscript{2}\hspace{0.02cm}Technische Universität Darmstadt\\[-3pt]
		{\tt\small \url{https://christophreich1996.github.io/differentiable_jpeg/}}
	}
	{
		\twocolumn[{
			\renewcommand\twocolumn[1][]{#1}
			\maketitle
			\begin{center}
				\sffamily
				\vspace{-1.0em}
				\begin{tabular}{c c c | c c c}
					\multicolumn{3}{c|}{\cellcolor{tud1a!20} JPEG quality = 50} & \multicolumn{3}{c}{\cellcolor{tud3c!20} JPEG quality = 1} \\\addlinespace[-.05em]
					\cellcolor{tud1a!20} {\scriptsize JPEG (OpenCV)} & 
					\cellcolor{tud1a!20} {\scriptsize Diff. JPEG (Shin \etal)} & 
					\cellcolor{tud1a!20} {\scriptsize Diff. JPEG (ours)} & 
					\cellcolor{tud3c!20} {\scriptsize JPEG (OpenCV)} & 
					\cellcolor{tud3c!20} {\scriptsize Diff. JPEG (Shin \etal)} &
					\cellcolor{tud3c!20} {\scriptsize Diff. JPEG (ours)} \\\addlinespace[-.02em]
					\cellcolor{tud1a!20} \includegraphics[width=0.1325\linewidth, trim={0 8.25cm 0 15cm}, clip]{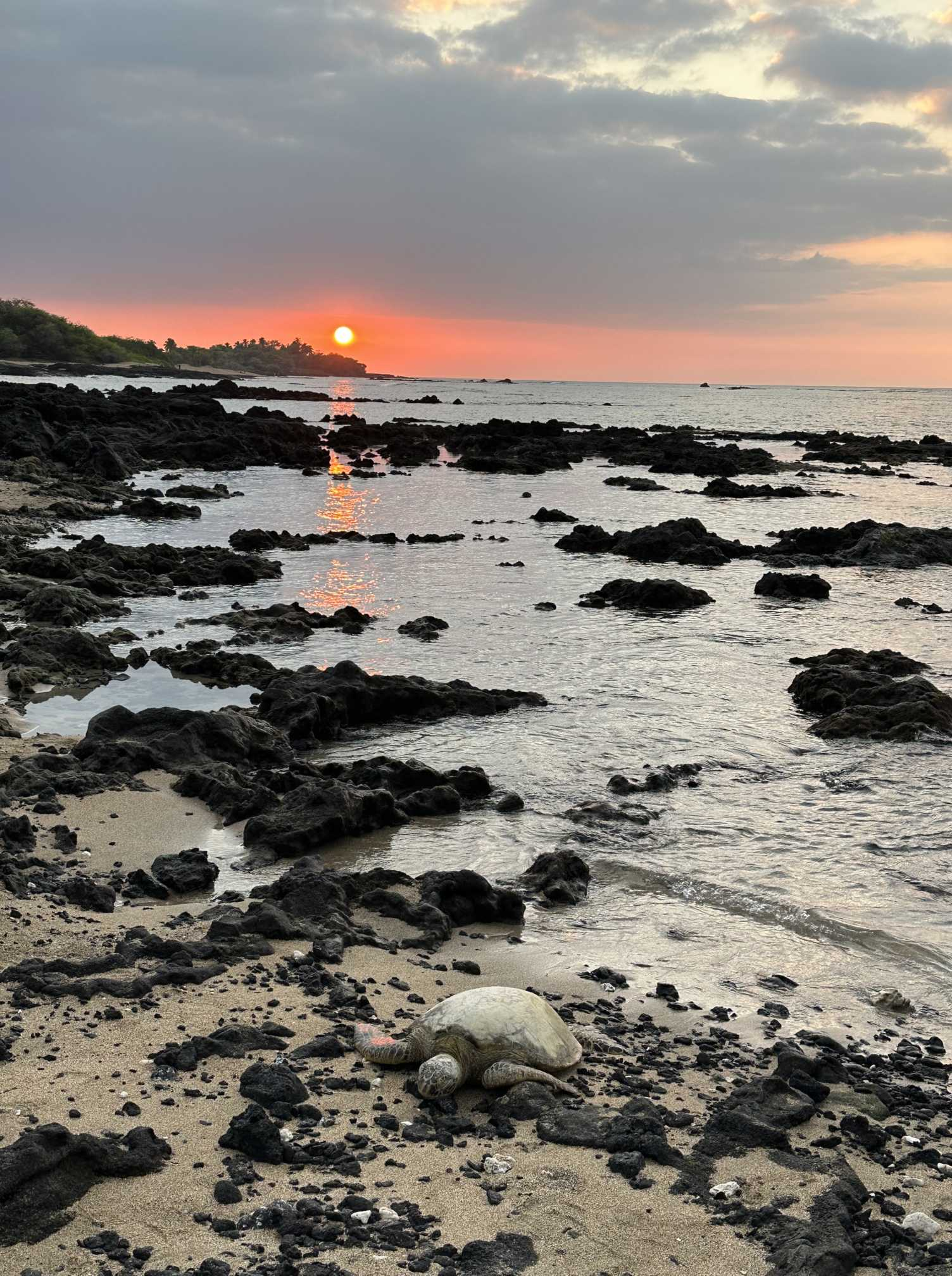} &
					\cellcolor{tud1a!20} \includegraphics[width=0.1325\linewidth, trim={0 8.25cm 0 15cm}, clip]{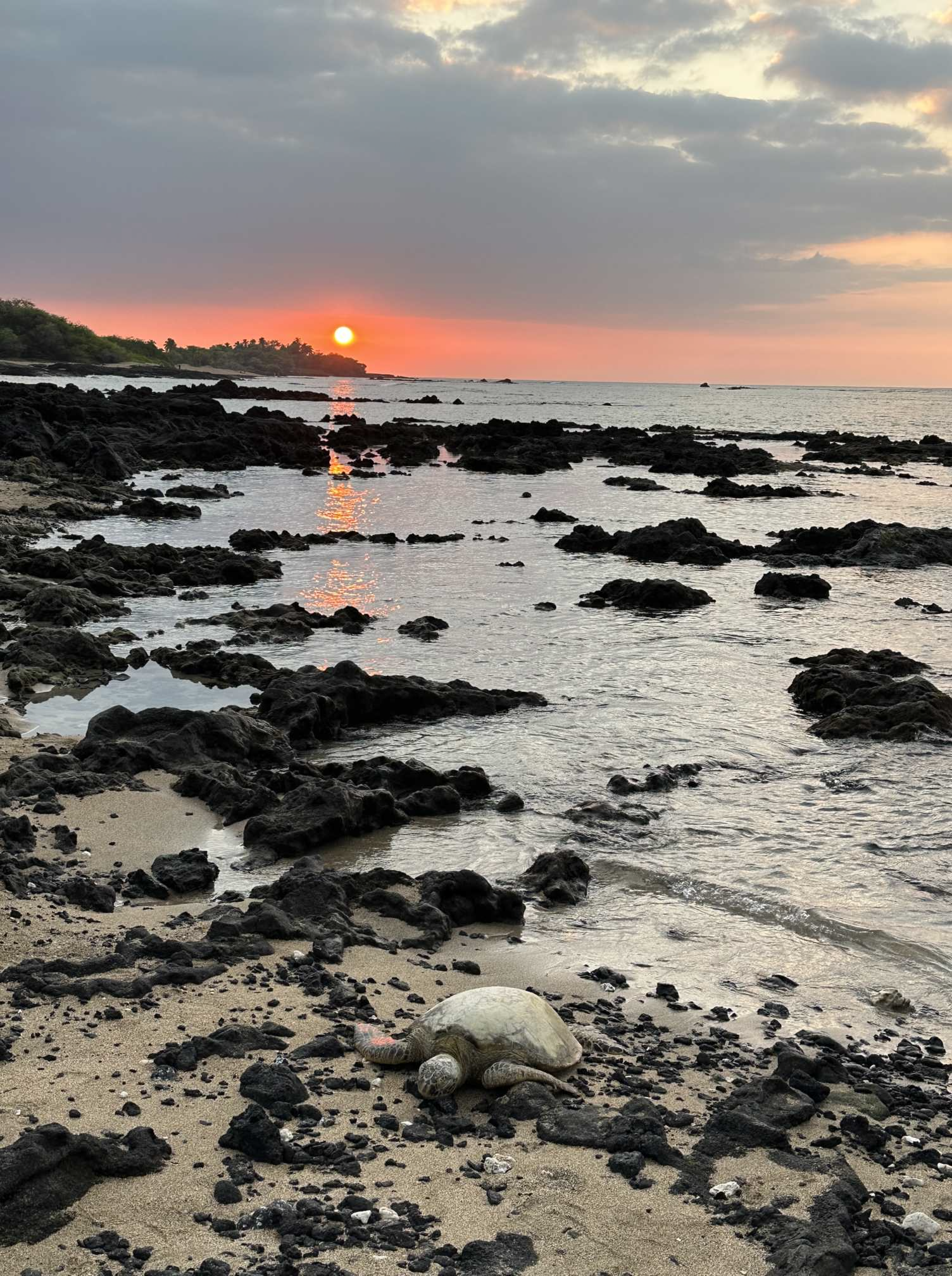} & 
					\cellcolor{tud1a!20} \includegraphics[width=0.1325\linewidth, trim={0 8.25cm 0 15cm}, clip]{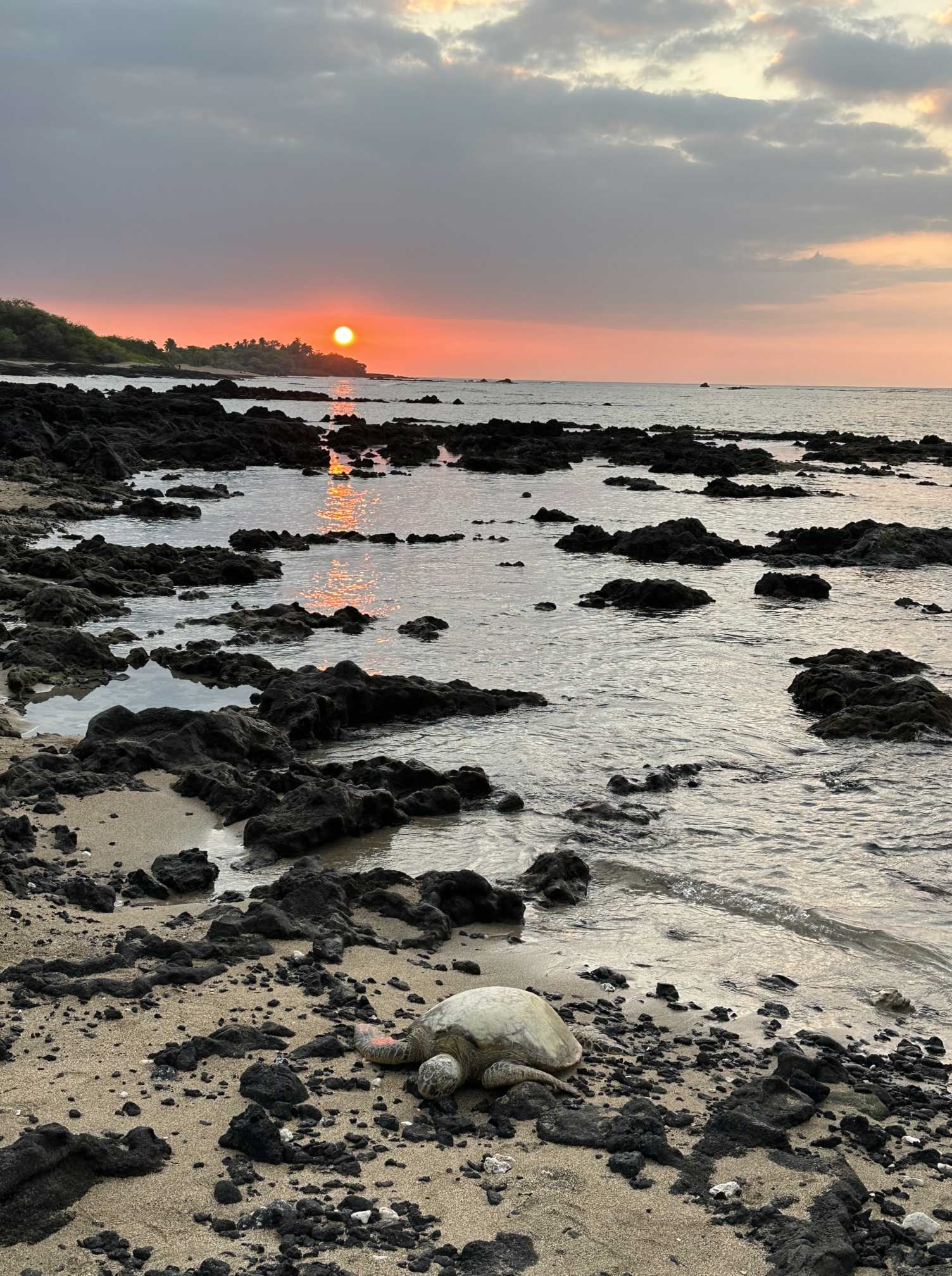} & 
					\cellcolor{tud3c!20} \includegraphics[width=0.1325\linewidth, trim={0 8.25cm 0 15cm}, clip]{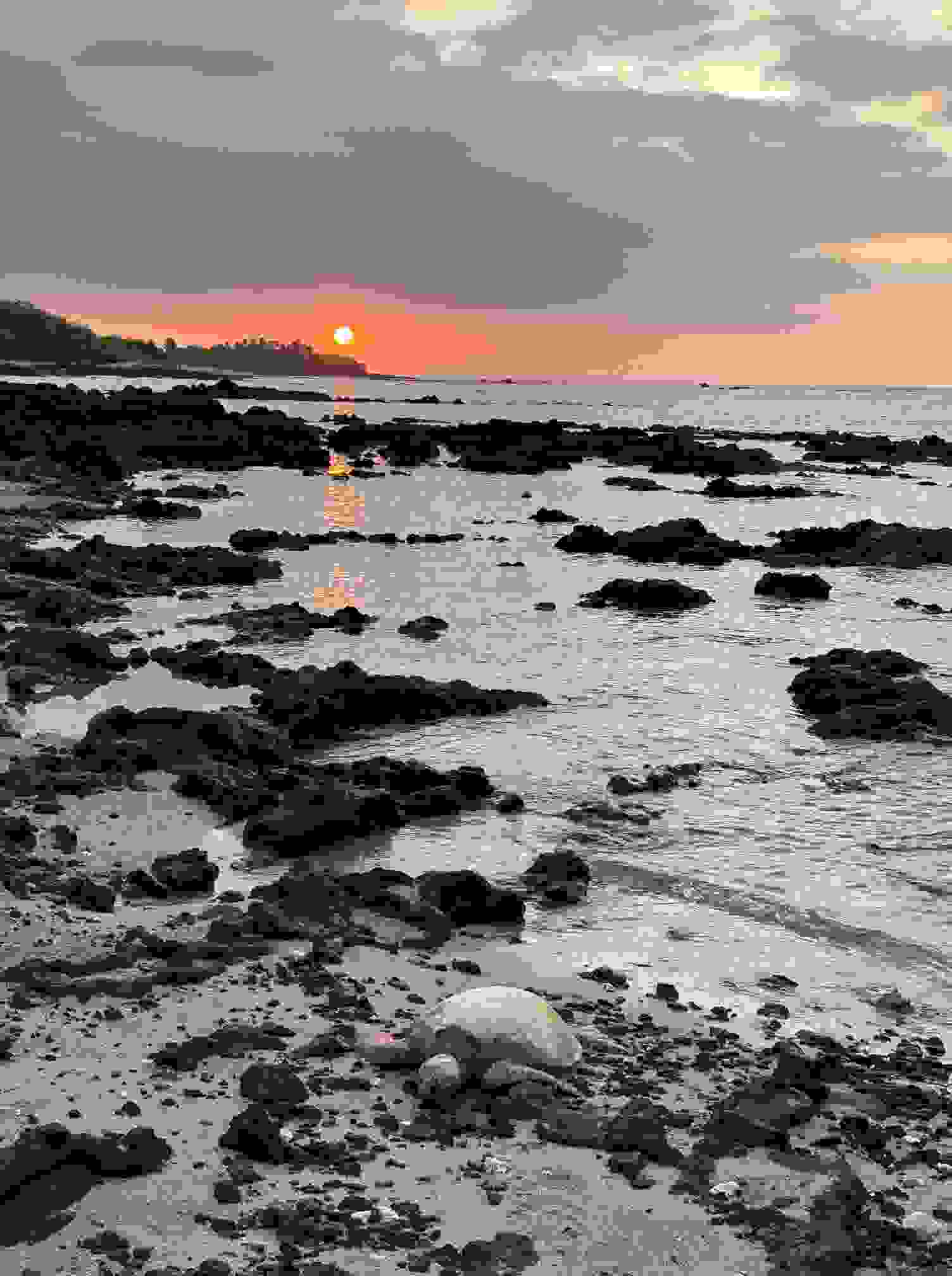} &
					\cellcolor{tud3c!20} \includegraphics[width=0.1325\linewidth, trim={0 8.25cm 0 15cm}, clip]{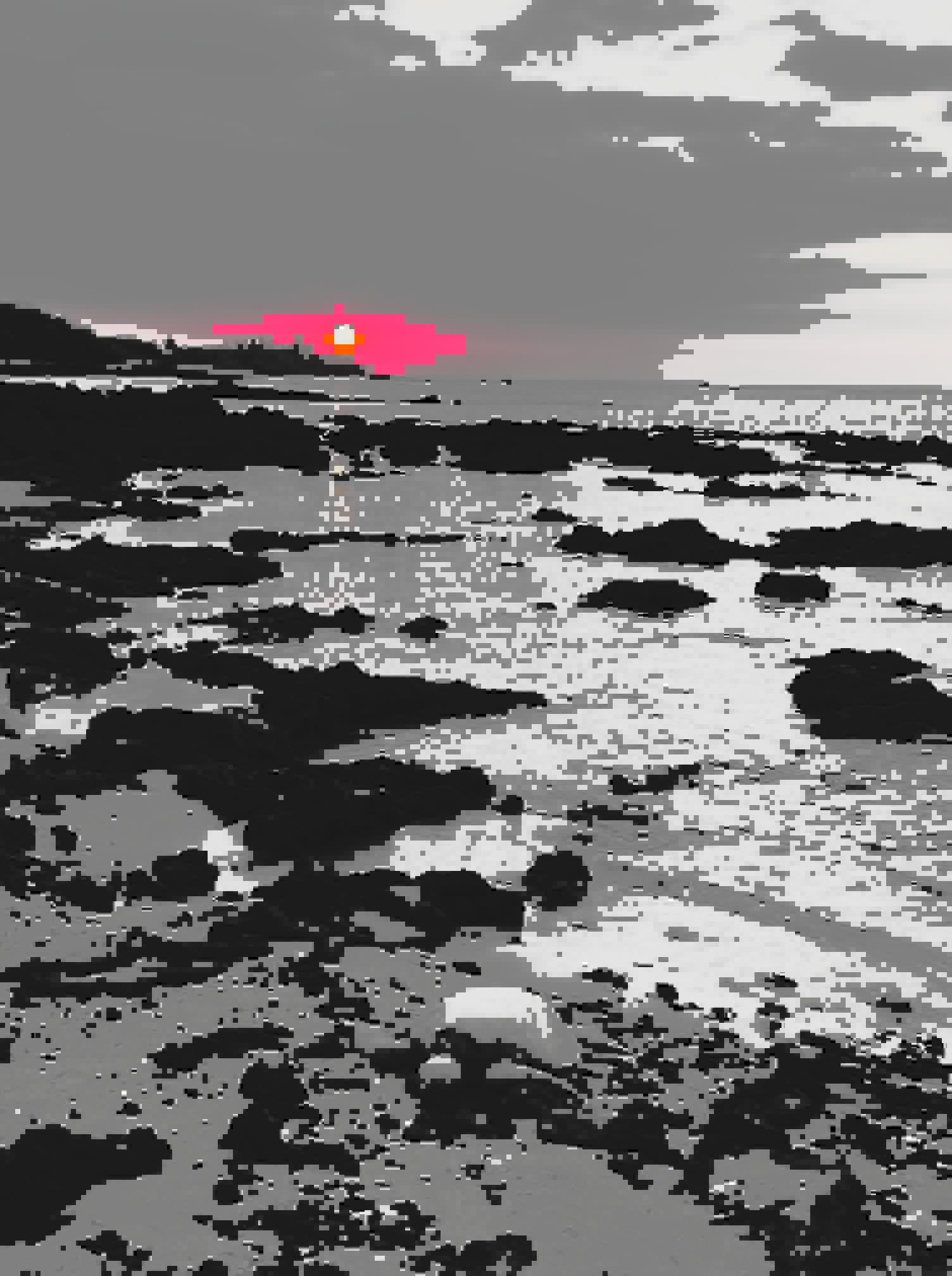} & 
					\cellcolor{tud3c!20} \includegraphics[width=0.1325\linewidth, trim={0 8.25cm 0 15cm}, clip]{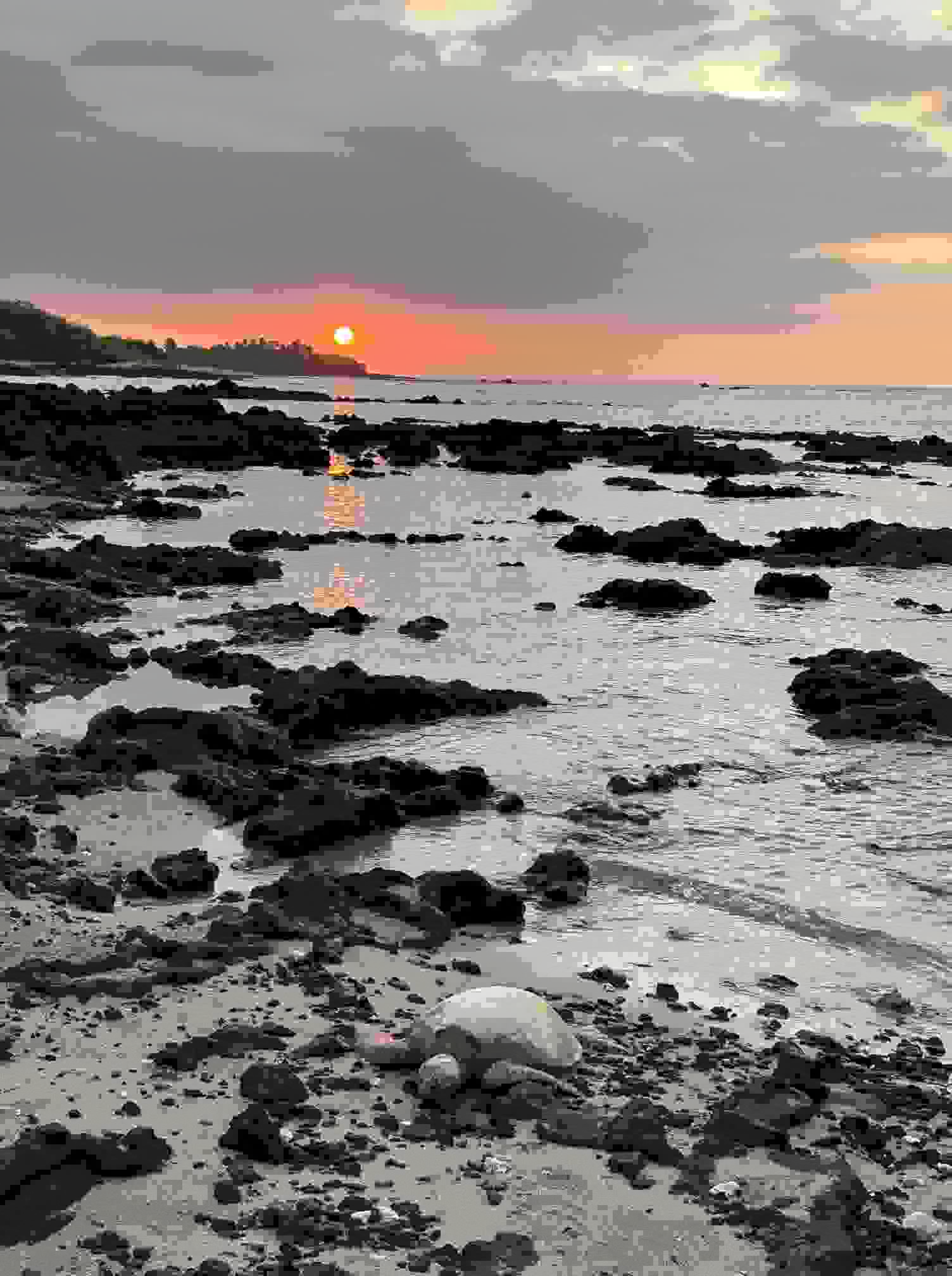} \\\addlinespace[-.325em]
					\cellcolor{tud1a!20} & 
					\cellcolor{tud1a!20} {\scriptsize PSNR = 43.13dB} & 
					\cellcolor{tud1a!20} {\scriptsize PSNR = 43.25dB} & 
					\cellcolor{tud3c!20} & 
					\cellcolor{tud3c!20} {\scriptsize PSNR = 18.05dB} & 
					\cellcolor{tud3c!20} {\scriptsize PSNR = 36.39dB} \\\addlinespace[-.325em]
					\multirow{-2}{*}{\cellcolor{tud1a!20} {\scriptsize Reference}} & 
					\cellcolor{tud1a!20} {\scriptsize SSIM = 0.996\,\,\,\,\,} & 
					\cellcolor{tud1a!20} {\scriptsize SSIM = 0.997\,\,\,\,\,} & 
					\multirow{-2}{*}{\cellcolor{tud3c!20} {\scriptsize Reference}} & 
					\cellcolor{tud3c!20} {\scriptsize SSIM = 0.616\,\,\,\,\,} & 
					\cellcolor{tud3c!20} {\scriptsize SSIM = 0.978\,\,\,\,\,} \\\addlinespace[-.325em]
				\end{tabular}
				\captionof{figure}{\textbf{Qualitative approximation results.} For a JPEG quality of 50, both Shin \etal~\cite{Shin2017} and our differentiable JPEG approach approximate the standard JPEG coding well. When reducing the JPEG quality to 1, the approach by Shin \etal does not approximate the JPEG coding well, while our differentiable JPEG still leads to a strong approximation. Structural similarity index measure (SSIM) and peak signal-to-noise ratio (PSNR) measured \wrt the coded image of the (non-differentiable) reference JPEG implementation (OpenCV~\cite{Bradski2000}).\label{fig:first}\\}
				\vspace{-0.95em}
			\end{center}
		}]
	}
	
	\thispagestyle{empty}
	%%%%%%%%% ABSTRACT
	\begin{abstract}
		JPEG remains one of the most widespread lossy image coding methods. However, the non-differentiable nature of JPEG restricts the application in deep learning pipelines. Several differentiable approximations of JPEG have recently been proposed to address this issue. This paper conducts a comprehensive review of existing diff. JPEG approaches and identifies critical details that have been missed by previous methods. To this end, we propose a novel diff. JPEG approach, overcoming previous limitations. Our approach is differentiable \wrt the input image, the JPEG quality, the quantization tables, and the color conversion parameters. We evaluate the forward and backward performance of our diff. JPEG approach against existing methods. Additionally, extensive ablations are performed to evaluate crucial design choices. Our proposed diff. JPEG resembles the (non-diff.) reference implementation best, significantly surpassing the recent-best diff. approach by \emph{$3.47$dB} (PSNR) on average. For strong compression rates, we can even improve PSNR by \emph{$9.51$dB}. Strong adversarial attack results are yielded by our diff. JPEG, demonstrating the effective gradient approximation. Our code is available at \url{https://github.com/necla-ml/Diff-JPEG}.
		
	\end{abstract}
	\thispagestyle{firststyle}
	
	% Add content
	\input{content/introduction.tex}
	\input{content/related_work.tex}
	\input{content/method.tex}
	\input{content/experiments.tex}
	\input{content/conclusion.tex}

	{\small
		\bibliographystyle{ieee_fullname}
		\bibliography{references}
	}
	
\end{document}

%% file: packages.tex
\usepackage{graphicx}
\usepackage{amsmath}
\usepackage{amsfonts}
\usepackage{amssymb}
\usepackage{amsthm}
\usepackage{booktabs}
\usepackage{pifont}
\usepackage{tabularx}
\usepackage{colortbl}
\usepackage{calc}
\usepackage{multirow}
\usepackage{makecell}
\usepackage{tikz}
\usepackage[export]{adjustbox}
\usepackage{pgfplots}
\usepackage{fontawesome}
\usepackage[per-mode=symbol, group-digits=false, detect-all=true,input-symbols={()}]{siunitx}
\usepackage{float}
\usepackage{listings}
\usepackage[inline]{enumitem}
\usepackage{algorithm}
\usepackage[format=plain,labelformat=simple,labelsep=period,font=small,skip=4pt,compatibility=false]{caption}
\usepackage[font=footnotesize,skip=2pt,subrefformat=parens]{subcaption}
\usetikzlibrary{spy}
\usetikzlibrary{shapes.arrows}
\usetikzlibrary{arrows.meta, arrows}
\usetikzlibrary{patterns,shapes.arrows}
\usetikzlibrary{positioning}
\usetikzlibrary{matrix}
\usetikzlibrary{pgfplots.groupplots}
\usepgfplotslibrary{groupplots,dateplot}
\usepgfplotslibrary{fillbetween}
\tikzset{every picture/.style={/utils/exec={\sffamily}}}
\pgfplotsset{compat=1.10}
\DeclareUnicodeCharacter{2212}{−}

%% file: colors.tex
\definecolor{tud0d}{cmyk/RGB/HTML}{0,0,0,.8/83,83,83/535353}
\definecolor{tud0c}{cmyk/RGB/HTML}{0,0,0,.6/137,137,137/898989}
\definecolor{tud0b}{cmyk/RGB/HTML}{0,0,0,.4/181,181,181/B5B5B5}
\definecolor{tud0a}{cmyk/RGB/HTML}{0,0,0,.2/220,220,220/DCDCDC}
\definecolor{tud1a}{cmyk/RGB/HTML}{.7,.4,0,0/93,133,195/5D85C3}
\definecolor{tud2a}{cmyk/RGB/HTML}{0.8,.2,0,0/0,156,218/009CDA}
\definecolor{tud3a}{cmyk/RGB/HTML}{0.7,0,.5,0/80,182,149/50B695}
\definecolor{tud4a}{cmyk/RGB/HTML}{.4,0,.8,0/175,204,80/AFCC50}
\definecolor{tud5a}{cmyk/RGB/HTML}{.2,0,.8,0/221,223,72/DDDF48}
\definecolor{tud6a}{cmyk/RGB/HTML}{0,.1,.7,0/255,224,92/FFE05C}
\definecolor{tud7a}{cmyk/RGB/HTML}{0,.3,.8,0/248,186,60/F8BA3C}
\definecolor{tud8a}{cmyk/RGB/HTML}{0,.6,.8,0 /238,122,52/EE7A34}
\definecolor{tud9a}{cmyk/RGB/HTML}{0,.8,.7,0/233,80,62/E9503E}
\definecolor{tud10a}{cmyk/RGB/HTML}{.2,.9,0,0/201,48,142/C9308E}
\definecolor{tud11a}{cmyk/RGB/HTML}{.6,.8,0,0/128,69,151/804597}
\definecolor{tud1b}{cmyk/RGB/HTML}{1,.6,0,0/0,90,169/005AA9}
\definecolor{tud2b}{cmyk/RGB/HTML}{1,.3,0,0/0,131,204/0083CC}
\definecolor{tud3b}{cmyk/RGB/HTML}{1,0,.6,0/0,157,129/009D81}
\definecolor{tud4b}{cmyk/RGB/HTML}{.5,0,1,0/153,192,0/99C000}
\definecolor{tud5b}{cmyk/RGB/HTML}{.3,0,1,0/201,212,0/C9D400}
\definecolor{tud6b}{cmyk/RGB/HTML}{0,.2,1,0/253,202,0/FDCA00}
\definecolor{tud7b}{cmyk/RGB/HTML}{0,.4,1,0/245,163,0/F5A300}
\definecolor{tud8b}{cmyk/RGB/HTML}{0,.7,1,0/236,101,0/EC6500}
\definecolor{tud9b}{cmyk/RGB/HTML}{0,1,.9,0/230,0,26/E6001A}
\definecolor{tud10b}{cmyk/RGB/HTML}{.4,1,0,0/166,0,132/A60084}
\definecolor{tud11b}{cmyk/RGB/HTML}{.7,1,0,0/114,16,133/721085}
\definecolor{tud1c}{cmyk/RGB/HTML}{1,.7,.2,0/0,78,138/004E8A}
\definecolor{tud2c}{cmyk/RGB/HTML}{1,.5,.2,0/0,104,157/00689D}
\definecolor{tud3c}{cmyk/RGB/HTML}{1,.2,.6,0/0,136,119/008877}
\definecolor{tud4c}{cmyk/RGB/HTML}{.6,.1,1,0/127,171,22/7FAB16}
\definecolor{tud5c}{cmyk/RGB/HTML}{.4,.1,1,0/177,189,0/B1BD00}
\definecolor{tud6c}{cmyk/RGB/HTML}{.2,.3,1,0/215,172,0/D7AC00}
\definecolor{tud7c}{cmyk/RGB/HTML}{.2,.5,1,0/210,135,0/D28700}
\definecolor{tud8c}{cmyk/RGB/HTML}{.2,.8,1,0/204,76,3/CC4C03}
\definecolor{tud9c}{cmyk/RGB/HTML}{.3,1,.9,0/185,15,34/B90F22}
\definecolor{tud10c}{cmyk/RGB/HTML}{.5,1,.3,0/149,17,105/951169}
\definecolor{tud11c}{cmyk/RGB/HTML}{.8,1,.2,0/97,28,115/611C73}
\definecolor{tud1d}{cmyk/RGB/HTML}{1,.9,.3,0/36,53,114/243572}
\definecolor{tud2d}{cmyk/RGB/HTML}{1,.7,.4,0/0,78,115/004E73}
\definecolor{tud3d}{cmyk/RGB/HTML}{1,.4,.7,0/0,113,94/00715E}
\definecolor{tud4d}{cmyk/RGB/HTML}{.7,.3,1,0/106,139,55/6A8B22}
\definecolor{tud5d}{cmyk/RGB/HTML}{.5,.2,1,0/153,166,4/99A604}
\definecolor{tud6d}{cmyk/RGB/HTML}{.4,.4,1,0/174,142,0/AE8E00}
\definecolor{tud7d}{cmyk/RGB/HTML}{.3,.6,1,0/190,111,0/BE6F00}
\definecolor{tud8d}{cmyk/RGB/HTML}{.4,.8,1,0/169,73,19/A94913}
\definecolor{tud9d}{cmyk/RGB/HTML}{.5,1,.9,0/156,28,38/961C26}
\definecolor{tud10d}{cmyk/RGB/HTML}{.7,1,.5,0/115,32,84/732054}
\definecolor{tud11d}{cmyk/RGB/HTML}{.9,1,.3,0/76,34,106/4C226A}

\definecolor{necgray}{HTML}{7d7d7d}
\definecolor{necblue}{HTML}{2c328f}
\definecolor{necbluedark}{HTML}{00285e}
\definecolor{necorange}{HTML}{eb6e00}
\definecolor{necyellow}{HTML}{fff484}
\definecolor{green}{HTML}{388a73}

\definecolor{qp51}{rgb}{0.5,0,0}
\definecolor{qp0}{rgb}{0,0,0.5}

\definecolor{tud3b6b1}{RGB}{63, 168, 97}
\definecolor{tud3b6b2}{RGB}{127, 180, 65}
\definecolor{tud3b6b3}{RGB}{190, 191, 32}

%% file: content/introduction.tex
\section{Introduction}
\label{sec:introduction}

JPEG (Joint Photographic Experts Group) coding is a standardized lossy compression approach for digital images~\cite{Hudson2018, Wallace1992}. As one of the most popular image coding standards for storing and transmitting image data, JPEG coding has become an integral part of various devices and programs. The acceptable rate-distortion performance paired with a strong compression efficiency strikes a delicate balance for many applications. JPEG's straightforward implementation and support for parallel computing further bolsters its popularity and makes JPEG coding a preferred choice in many image processing pipelines~\cite{Hudson2018}.

The widespread use of JPEG in image processing pipelines has motivated the integration of JPEG coding into deep learning pipelines. Applications of JPEG coding in deep learning pipelines include (differentiable) data augmentation~\cite{Hataya2020, Jung2020, Karras2020, Shi2020, Shorten2019, Shu2021, Zhao2020}, data hiding~\cite{Zhang2020, Zhu2018}, deepfake detection~\cite{Yang2021}, adversarial attacks~\cite{Guo2018, Shin2017}, or optimizing JPEG for deep neural networks~\cite{Choi2020, Luo2021, Xie2022}. Using JPEG in deep learning pipelines requires non-zero gradients to be propagated through the JPEG encoding-decoding. However, due to its inherently discrete nature, JPEG encoding-decoding is non-differentiable. Motivated by this, considerable effort has been devoted to building various differentiable JPEG approaches~\cite{Choi2020, Luo2021, Shin2017, Strumpler2020, Xie2022, Yang2021, Zhang2020, Zhu2018}.

While various differentiable JPEG approaches have been proposed, we are not aware of any work providing a comparison of these approaches. In this paper, we conduct a comprehensive review of existing differentiable JPEG approaches and highlight crucial deficiencies (\eg, not considering discretizations of standard JPEG) and suboptimal design choices (\eg, poor rounding approximations) that impede precise predictions (\cf \cref{fig:first}) and effective gradients. To remedy the outlined deficiencies, we present a novel differentiable JPEG approach. While drawing inspiration from prior differentiable JPEG implementations, our differentiable JPEG approach makes use of novel components, such as differentiable clipping, and is the first to model all important details of standard (non-differentiable) JPEG. In addition, we also propose a straight-through estimator (STE) variant of our differentiable JPEG.

We thoroughly evaluate the performance of existing differentiable JPEG implementations and our proposed approach, in approximating standard (non-diff.) JPEG coding. We show that all existing diff. approaches fail to accurately resemble standard JPEG over the whole compression range (\cf \cref{fig:first}). To the best of our knowledge, our differentiable JPEG approach (w/ \& w/o STE) is the first to provide an accurate approximation of standard JPEG across the entire range of compression strengths, while offering gradients \wrt all inputs. This is qualitatively showcased in \cref{fig:first}. We validate the effectiveness of gradients derived from existing approaches and our differentiable JPEG by conducting adversarial attack experiments. Our approach generates superior adversarial samples in comparison to existing methods. These findings indicate that gradients obtained through our differentiable JPEG are notably more effective for gradient-based optimization tasks (\eg, neural network training) than those derived from existing methods.

%% file: content/related_work.tex
{
\begin{figure*}
    \centering
    \input{artwork/jpeg}
    \caption{\textbf{The JPEG encoding-decoding pipeline.} The original input image is encoded to a JPEG file in a lossy manner. To recover the coded image the encoding is reversed in the decoding. JPEG uses lossless coding in conjunction with lossy coding. Since no information is lost (identity mapping) during the lossless encoding/decoding we can neglect these coding steps in our differentiable JPEG approach.}
    \label{fig:jpeg}
    \vspace{-0.6em}
\end{figure*}
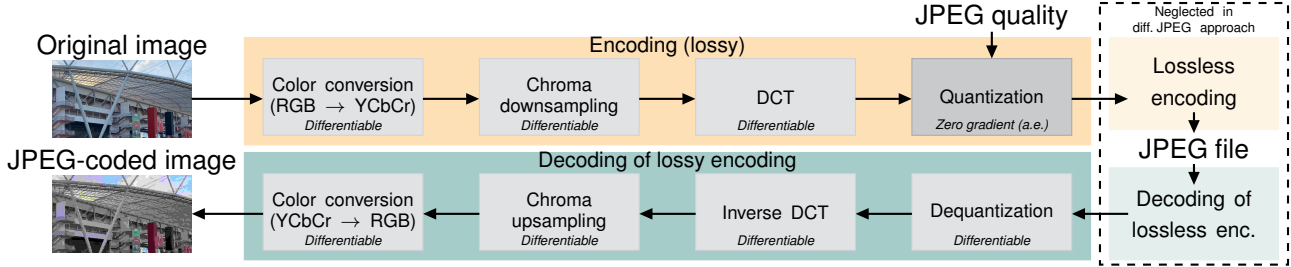
}

\section{Background: The JPEG Coding Standard}
\label{sec:jpegstandard}

The JPEG compression standard~\cite{Hussain2018, Wallace1992}, in the baseline mode, uses both lossy and lossless coding to achieve efficient image compression. The encoding starts by converting the original RGB image to the YCbCr color space and performing chroma downsampling. The YCbCr channels are then transformed into the frequency domain using a patch-wise discrete cosine transform (DCT). A given JPEG quality controls the quantization strength of the DCT features, trading file-size against distortion (\cf \cref{subsec:rate_distortion}). Finally, the compressed JPEG file is produced using lossless coding. During decoding, the lossless and lossy encoding steps are reversed to reconstruct the JPEG-coded image from the JPEG file. \cref{fig:jpeg} illustrates the JPEG coding process.

In general, JPEG encoding-decoding can be seen as a function mapping from an original (raw) RGB image $\mathbf{I}$ and the JPEG quality $\q$ to the JPEG-coded (distorted) image $\hat{\mathbf{I}}$
\begin{equation}\label{eq:jpeg_coding}
     \begin{aligned}
        \jpeg\left(\mathbf{I}, \q\right)&=\hat{\mathbf{I}},\;\;\q\in\{1, 2, \ldots, 99\}\\[-3pt]
        \mathbf{I},\hat{\mathbf{I}}&\in\{0, \ldots, 255\}^{3\times H\times W}.
    \end{aligned}
\end{equation}
$H$ and $W$ denote the image resolution. Some implementations consider a max. $\q$ of $100$, others of $99$, for the sake of generality we use $99$ as max. $\q$. In the next subsections, we describe the internals of the JPEG function in detail.

\subsection{JPEG encoding} \label{subsec:encoding}

The JPEG encoding process compresses a given image to a binary JPEG file and is composed of four main steps followed by lossless encoding (\cf \cref{fig:jpeg}). In the following, we explain the details of all encoding steps.

\paragraph{Color conversion (RGB $\to$ YCbCr).} Digital imagery is typically displayed using the RGB color space. JPEG makes use of the YCbCr color space for compression. To this end, JPEG converts the RGB image to the YCbCr color space by a pixel-wise affine transformation.

\paragraph{Chroma subsampling.} The human eye tends to be more sensitive to variations in brightness than to color details~\cite{Macadam1942}. This motivates the use of chroma subsampling in JPEG. By discarding less relevant information to the human eye, chroma subsampling introduces a minimal loss in perceptual quality while leading to compression. Chroma subsampling is typically implemented by an anti-aliasing operation (\eg, 2D convolution) followed by standard downsampling and is applied to both chroma channels (Cb \& Cr).

\paragraph{Patch-wise discrete cosine transform.} JPEG compression utilizes a patch-wise (and channel-wise) DCT-II operation to transform an image into a frequency (DCT) space. Before applying the DCT, non-overlapping $8\times 8$ patches from the chroma-subsampled YCbCr image are extracted. For a given (flatten) patch $\mathbf{p}\in\{0, 1, \ldots, 255\}^{64}$, the DCT is described by $\hat{\mathbf{p}}=\mathbf{a}\odot\mathbf{G}\mathbf{p}$. $\odot$ denotes the Hadamard product, $\mathbf{G}\in\mathbb{R}^{64\times 64}$ contains the DCT coefficients, and $\mathbf{a}$ is a scaling factor. $\mathbf{G}$ is computed by $G_{8u+v,8i+j}=\cos\!\left(\frac{2x + 1}{16}\right)\cos\!\left(\frac{2y+1}{16}\right)$ and $\mathbf{a}$ by $a_{8u+v}=\frac{1}{4}\alpha\!\left(u\right)\alpha\!\left(v\right)$ with $\alpha\!\left(u\right)=\begin{cases} \frac{1}{\sqrt{2}} & \text{if}\;u=0 \\[-1.5pt] 1 & \text{otherwise} \end{cases}$ and $u,v, i, j\in\{0, 1, \ldots, 7\}$. $\hat{\mathbf{p}}\in\mathbb{R}^{64}$ represents the transformed patch. For simplicity, we omit the channel (YCbCr) and patch indexing.

\paragraph{Quantization.} Through quantization, controlled by the JPEG quality $\q$, frequencies are suppressed for the sake of compression. During the quantization step, the given JPEG quality $\q$ is mapped to a scale factor $s$ by:
\begin{equation}\label{eq:qt_scale}
    s\!\left(\q\right)=\begin{cases} 
        \frac{5000}{\q} & \text{if}\;\q<50 \\[-1.5pt] 
        200-2\q & \text{otherwise}.
    \end{cases}
\end{equation}
The scale factor is applied to the (standard) quantization table $\qt_{\text{s}}\in\{1, 255\}^{8\times 8}$ by ${\hat{\qt}=\frac{s\,\qt_{\text{s}} + 50}{100}}$. The scaled quantization table is applied to each 2D DCT patch $\hat{\mathbf{P}}\in\mathbb{R}^{8\times 8}$ (reshaped $\hat{\mathbf{p}}$) followed by the application of the rounding function $\overline{P}_{m,n}=\left\lfloor \frac{\hat{P}_{m,n}}{\hat{QT}_{m,n}}\right\rceil$ with $m,n\in\{0, 1,\ldots, 7\}$. $\lfloor\cdot\rceil$ denotes the rounding to the next integer. Note that standard JPEG performs integer arithmetic to compute $s$ and $\hat{\qt}$ this is equivalent to applying the floor function $\lfloor\cdot\rfloor$. Additionally, $\hat{\qt}$ is clipped to the integer range of $\{1, 2,\ldots, 255\}$. Note JPEG also supports custom quantization tables and uses two separate tables for the luma channel (Y) as well as the chroma channels (Cb \& Cr). For simplicity, we do not distinguish between quantization tables.

\paragraph{Lossless encoding.} JPEG utilizes lossless entropy coding to compress all quantized DCT patches $\hat{\mathbf{P}}$. The lossless encoding first arranges the lossy encoded patches in a zigzag order before performing run-length encoding. Finally, Huffman coding is performed to build the binary JPEG file. Note the final JPEG file includes not only the encoded image content but also the scaled quantization tables and other markers including information such as the image resolution.

\subsection{JPEG decoding} \label{subsec:decoding}

The JPEG decoding converts the compressed binary JPEG file back to an RGB image (\cf \cref{fig:jpeg}). Four main steps and the inversion of the lossless encoding operations compose the decoding. On a high level, every decoding step reverses the corresponding encoding step (\cf \cref{fig:jpeg}).

\paragraph{Decoding of lossless encoding.} The Huffman-encoded JPEG file is decoded before the run-length encoding is undone. Finally, the information is rearranged as a pixel grid with three channels. Note that lossless encoding and decoding can be viewed as an identity mapping.

\paragraph{Dequantization.} To dequantize, the quantized DCT features are multiplied with the respective scaled QT (luma or chroma table) $\tilde{\mathbf{P}}=\overline{\mathbf{P}}\odot\hat{\qt}$.

\paragraph{Inverse patch-wise discrete cosine transform.} To convert the DCT information back into pixel space, the inverse discrete cosine transform is applied to each $8\times8$ patch.

\paragraph{Chroma upsampling.} To recover the original image resolution, both chroma channels are upsampled using bilinear interpolation.

\paragraph{Color conversion (YCbCr $\to$ RGB).} The coded image (YCbCr) is converted back into the RGB color space by applying the inverse of the previous affine transformation.

\subsection{JPEG rate-distortion trade-off} \label{subsec:rate_distortion}

\begin{figure}[t]
    \centering
    \begin{subfigure}{.285\linewidth}
        \centering
        \includegraphics[width=\linewidth]{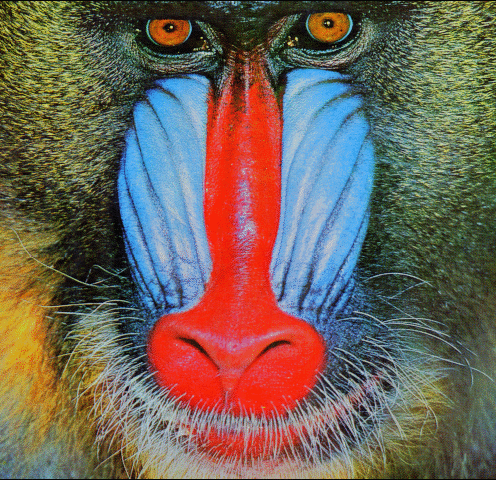}
        \caption{Original image}
        \label{subfig:inputimage}
    \end{subfigure}\hfill%
    \begin{subfigure}{.285\linewidth}
        \centering
        \includegraphics[width=\linewidth]{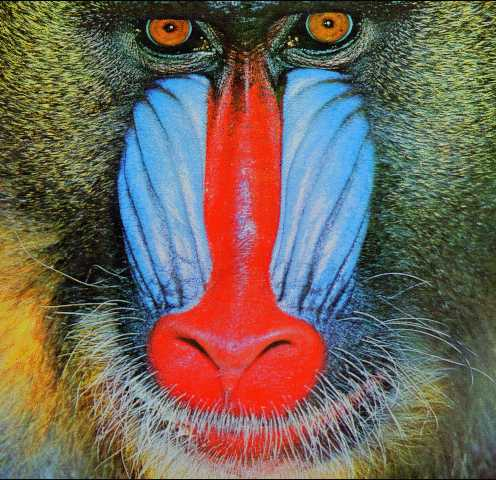}
        \caption{JPEG quality $50$}
        \label{subfig:jpegimage50}
    \end{subfigure}\hfill%
    \begin{subfigure}{.285\linewidth}
        \centering
        \includegraphics[width=\linewidth]{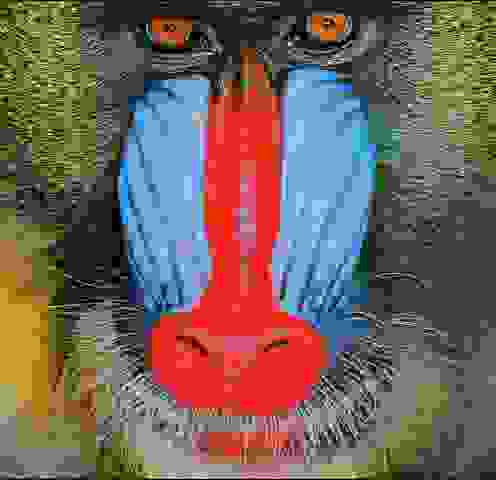}
        \caption{JPEG quality $1$}
        \label{subfig:jpegimage1}
    \end{subfigure}%
    \caption{\textbf{JPEG coding artifacts.} \emph{\subref{subfig:inputimage}} Original image, \emph{\subref{subfig:jpegimage50}} JPEG-coded image with a JPEG quality of $50$, file size is $47.3\si{\kilo\byte}$, and \emph{\subref{subfig:jpegimage1}} coded image with a JPEG quality of $1$, file size is $6.2\si{\kilo\byte}$. Image from the Set14~\cite{Zeyde2012} and OpenCV~\cite{Bradski2000} JPEG used.}
    \label{fig:jpeg_example}
    \vspace{-0.6em}
\end{figure}

JPEG has strong support for different compression strengths. By adjusting the JPEG quality parameter $\q$, different compression strengths can be achieved. A low JPEG quality results in a small file size but leads to significant image distortion (\cf \cref{subfig:jpegimage1}) since quantization suppresses plenty of frequencies. \textit{Vice versa}, a high JPEG quality leads to a larger file size but reduces distortion (\cf \cref{subfig:jpegimage50}). This tradeoff is known as the rate-distortion trade-off.

JPEG employs the DCT in a patch-wise manner to achieve efficient compression. This compression approach introduces distinctive artifacts in the resulting distorted JPEG-coded image. These artifacts manifest in various forms, including ringing, contouring, posterizing, and most notably, block boundary artifacts. Among these artifacts, block boundary artifacts are particularly noticeable when compressing natural images (\cf \cref{subfig:jpegimage50}). 

\subsection{Non-differentiability of JPEG}

The JPEG encoding-decoding process (\cf \cref{eq:jpeg_coding}) is an inherently discrete operation that precludes the application of continuous differentiation. However, we can extend all operations of the JPEG encoding-decoding to real-valued numbers. Consequently, we can formulate a continuous JPEG function, $\jpeg_c$, that accepts continuous inputs in terms of the original image and JPEG quality and produces a continuous coded image (${\jpeg_c:\mathbb{R}^{3\times H\times W}\times[1, 99]\to\mathbb{R}^{3\times H\times W}}$). This naive continuous generalization of the JPEG encoding-decoding suffers from a major limitation. The gradient of $\jpeg_c$ (\wrt to both inputs) is zero almost everywhere (a.e.) and undefined at points of jump discontinuity. This is caused by the reliance on rounding and floor functions in the encoding process. This property inhibits the direct integration of $\jpeg_c$ into gradient-based learning systems (\eg, deep neural network training) as they rely on the availability of ``useful'' non-zero gradients for optimization.

\section{Existing Differentiable JPEG Approaches}
\label{sec:relatedwork}

Existing differentiable JPEG approaches can be broadly categorized into three groups: straight-through estimator approaches, surrogate model approaches, and noise-based methods. All approaches aim to propagate ``useful'' gradients through the JPEG encoding-decoding.

\paragraph{STE approaches.} In standard STE, the gradient of a non-differentiable function is approximated by assuming a constant gradient of one~\cite{Bengio2013}. During the forward pass, the true non-differentiable function is used. STE has been shown to be effective in various deep learning-based approaches~\cite{Bengio2013, Esser2021, Jang2017, Papamakarios2021, Van2017}. This technique is also used when the gradient of a function would be zero a.e. (\eg, rounding function). Choi \etal~\cite{Choi2020} used STE to propagate gradients through the rounding operation of the JPEG encoding~\cite{Theis2017}. Instead of assuming a constant gradient as in standard STE, Xie \etal~\cite{Xie2022} utilizes the gradient of a tanh-based differentiable rounding approximation in the backward pass. Both approaches do not model the JPEG quality scaling of the quantization tables, limiting the general usability beyond the application proposed in the respective papers. Additionally, the bounded nature of the quantization tables and the quantization table scale is not considered. Nor is the bounded nature of the coded image modeled.

\paragraph{Surrogate model approaches.} Another technique to achieve ``useful'' gradients is to replace all non-differentiable components of JPEG coding (\eg, rounding) with differentiable approximations. The resulting differentiable JPEG approach is both an approximation in the forward and backward pass. Shin \etal proposed the first differentiable surrogate of the JPEG encoding-decoding~\cite{Shin2017}. A polynomial approximation of the rounding function ${\lfloor x\rceil+\left(x - \lfloor x\rceil\right)^3}$ enables the propagation of gradient through the rounding operation. Additionally, the quantization table scaling by the JPEG quality $\q$ is reformulated to \linebreak $s\!\left(\q\right)=\begin{cases} \frac{50}{\q} & \text{if}\;\q<50 \\[-1.5pt] 2-\frac{2\q}{100} & \text{otherwise} \end{cases}$ and $\hat{\qt}=s\qt_{\text{s}}$. Note this leads to a difference of $0.5$ in the scale quantization table $\hat{\qt}$ compared to the standard scaling factor computation (\cf \cref{eq:qt_scale}), subsequently deteriorating the approximation performance. Other approaches have been built on the approach by Shin \etal with slight modifications~\cite{Luo2021, Strumpler2020, Xing2021}. Instead of a polynomial approximation, Xing \etal approximate the rounding function with finite Fourier series approximation $x-\sum^{10}_{k=1}\!\!\frac{(-1)^{k+1}}{k\pi}\!\sin(2\pi kx)$. While Shin \etal does not model integer divisions nor the bounded nature of the quantization tables as well as the coded image, Xing \etal hard clips the coded image to the valid pixel range, leading to a zero-valued gradient for clipped pixels.

\paragraph{Noise-based approaches.} JPEG coding introduces unique distortion artifacts to the coded image (\cf \cref{subsec:rate_distortion} and \cref{fig:jpeg_example}). This motivates noise-based approaches, wherein JPEG coding is approximated by introducing specific noise into the original image. Zhu \etal~\cite{Zhu2018} achieves this by randomly applying dropout to the DCT features mimicking JPEG distortion. However, applying random dropout offers very limited control over the precise quality, leads to sparse gradients, and is only a coarse and stochastic approximation. Zhang \etal~\cite{Zhang2020} adds the true distortion as pseudo noise to the original image. While the resulting coded images match the true coded image, this approach is equivalent to applying STE to the full JPEG coding function. The resulting gradients are fully independent of the JPEG function. Additionally, both approaches only offer gradients \wrt the original image, severely limiting general applicability. Due to these major limitations, we do not consider noise-based approach as differentiable approximations.

\paragraph{JPEG file size modeling.} Certain applications require a differentiable estimate of the JPEG file size~\cite{Luo2021, Xie2022}. As all differentiable JPEG approaches neglect the lossless encoding/decoding, modeling the file size is non-trivial~\cite{Luo2021, Shin2017, Xie2022, Xing2021}. Existing approaches typically train a deep neural network to regress the file size from the quantized DCT features of the differentiable JPEG approach~\cite{Luo2021, Xie2022}. Note that the scope of this paper is to model JPEG encoding-decoding in a differentiable manner and not to model the JPEG file size. We refer the reader to Luo \etal~\cite{Luo2021} and Xie \etal~\cite{Xie2022} for more details on file-size modeling.

%% file: artwork/jpeg.tex
% \documentclass[crop,tikz]{standalone}

% \usepackage{tikz}
% \usepackage{graphicx}
% \usetikzlibrary{shapes.arrows}
% \usetikzlibrary{arrows.meta, arrows}
% \usetikzlibrary{patterns,shapes.arrows}
% \input{colors}

% \begin{document}
\scalebox{1.0}{
\begin{tikzpicture}[xscale=1, yscale=0.51, every node/.style={scale=1.0}, >={Stealth[inset=0pt,length=5pt,angle'=45]}]

    % Input image
    \begin{scope}[xshift=0.125cm, yshift=0cm]
        \node[] at (0, 4.35) {Original image};
        \node[anchor=center] at (0, 3) {\includegraphics[width=1.85cm, trim={0 5.5cm 0 5.5cm}, clip]{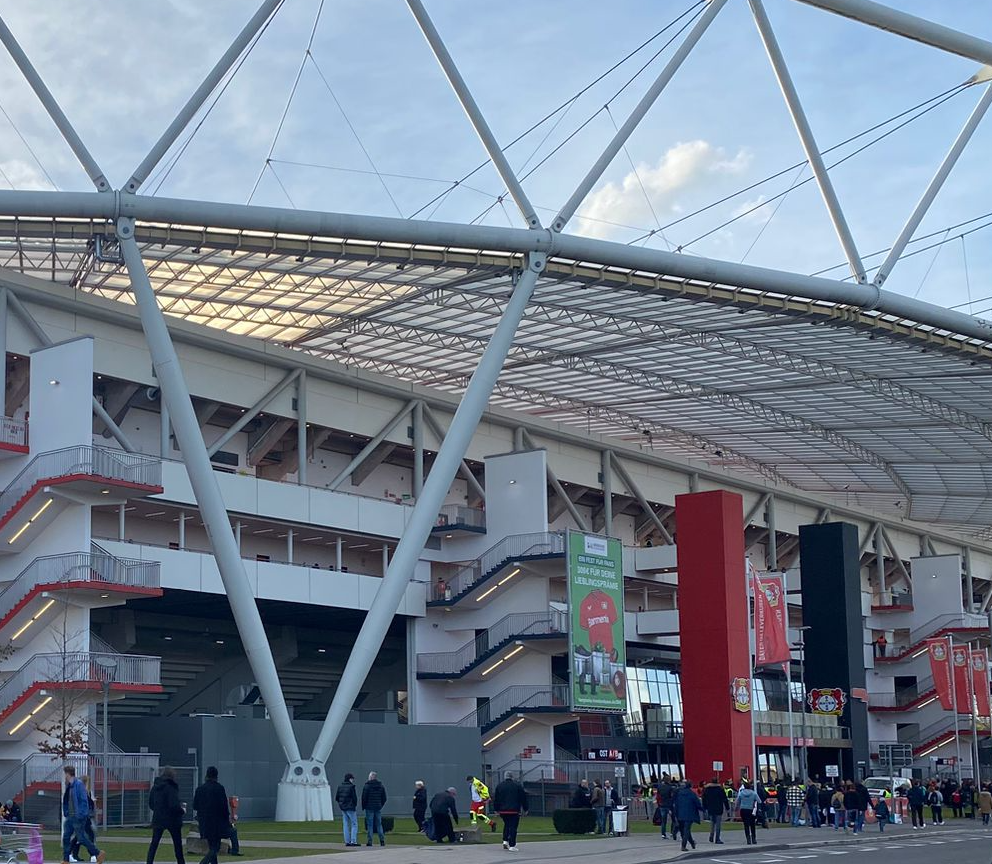}};
    \end{scope}
    
    % Draw encoding background
    \draw[tud7b!30, fill=tud7b!30] (1.75, 4.6) rectangle (13, 1.8);
    \node[] at (7.375, 4.35) {\footnotesize Encoding (lossy)};
    
    % Color conversion
    \draw[thick, ->] (1.05, 3) -- (2.0, 3);
    \draw[tud0a, fill=tud0a!60] (2.0, 4.05) rectangle (4.125, 2.05) node[pos=.5, text width=2cm, black, align=center] {\scriptsize Color conversion\\[-4pt] (RGB $\to$ YCbCr)};
    \node[] at (3.0625, 2.3) {\tiny \textit{Differentiable}};
    \draw[thick, ->] (4.125, 3) -- (4.875, 3);

    % Chroma downsampling
    \draw[tud0a, fill=tud0a!60] (4.875, 4.05) rectangle (7.0, 2.05) node[pos=.5, text width=2cm, black, align=center] {\scriptsize Chroma\\[-4pt] downsampling};
    \node[] at (5.9375, 2.3) {\tiny \textit{Differentiable}};
    \draw[thick, ->] (7.0, 3) -- (7.75, 3);

    % DCT
    \draw[tud0a, fill=tud0a!60] (7.75, 4.05) rectangle (9.875, 2.05) node[pos=.5, text width=2cm, black, align=center] {\scriptsize DCT};
    \node[] at (8.8125, 2.3) {\tiny \textit{Differentiable}};
    \draw[thick, ->] (9.875, 3) -- (10.625, 3);
    
    % Quantization
    \draw[tud0b, fill=tud0b!60] (10.625, 4.05) rectangle (12.75, 2.05) node[pos=.5, text width=2cm, black, align=center] {\scriptsize Quantization};
    \node[] at (11.6875, 2.3) {\tiny \textit{Zero gradient (a.e.)}};
    \draw[thick, <-] (11.6875, 4.05) -- (11.6875, 4.75) node[above=-3pt] {JPEG quality};

    % Box around lossless encoding and decoding
    \draw[dashed, thick] (13.125, 5.5) -- node[midway, below=-2.5pt, text width=2.5cm, align=center] {\tiny Neglected in\\[-6pt] diff.$\,$JPEG approach} (15.625, 5.5) -- (15.625, -1.325) -- (13.125, -1.325) -- cycle;
    
    % Draw lossless encoding background
    \draw[tud7b!10, fill=tud7b!10] (13.25, 4.6) rectangle (15.5, 2.2) node[pos=.5, text width=2.25cm, black, align=center] {\footnotesize Lossless\\ encoding};
    \draw[thick, ->] (12.75, 3) -- (13.5, 3);
    \draw[thick, ->] (14.375, 2.65) -- (14.375, 2.1);
    
    % Compressed object
    \node[] at (14.375, 1.7) {JPEG file};
    
    % Draw lossless decoding background
    \draw[tud3c!10, fill=tud3c!10] (13.25, 1.2) rectangle (15.5, -1.2) node[pos=.5, text width=2.25cm, black, align=center] {\footnotesize Decoding of\\ lossless enc.};
    \draw[thick, ->] (14.375, 1.3) -- (14.375, 0.75);
    
    % Draw decoding background
    \draw[tud3c!35, fill=tud3c!35] (1.75, 1.6) rectangle (13, -1.2);
    \node[] at (7.375, 1.35) {\footnotesize Decoding of lossy encoding};
    \draw[thick, ->] (13.5, 0) -- (12.75, 0);

    \begin{scope}[xshift=0cm, yshift=-3cm]
        % Dequantization
        \draw[tud0a, fill=tud0a!60] (10.625, 4.05) rectangle (12.75, 2.05) node[pos=.5, text width=2cm, black, align=center] {\scriptsize Dequantization};
        \node[] at (11.6875, 2.3) {\tiny \textit{Differentiable}};

        % Inverse DCT
        \draw[tud0a, fill=tud0a!60] (7.75, 4.05) rectangle (9.875, 2.05) node[pos=.5, text width=2cm, black, align=center] {\scriptsize Inverse DCT};
        \node[] at (8.8125, 2.3) {\tiny \textit{Differentiable}};
        \draw[thick, <-] (9.875, 3) -- (10.625, 3);

        % Chroma upsampling
        \draw[tud0a, fill=tud0a!60] (4.875, 4.05) rectangle (7.0, 2.05) node[pos=.5, text width=2cm, black, align=center] {\scriptsize Chroma\\[-4pt] upsampling};
        \node[] at (5.9375, 2.3) {\tiny \textit{Differentiable}};
        \draw[thick, <-] (7.0, 3) -- (7.75, 3);

        % Color conversion
        \draw[tud0a, fill=tud0a!60] (2.0, 4.05) rectangle (4.125, 2.05) node[pos=.5, text width=2cm, black, align=center] {\scriptsize Color conversion\\[-4pt] (YCbCr $\to$ RGB)};
        \node[] at (3.0625, 2.3) {\tiny \textit{Differentiable}};
        \draw[thick, <-] (4.125, 3) -- (4.875, 3);
    \end{scope}
    
    % Output image
    \begin{scope}[xshift=0.125cm, yshift=0cm]
        \node[] at (0, 1.35) {JPEG-coded image};
        \node[anchor=center] at (0, 0) {\includegraphics[width=1.85cm, trim={0 5.5cm 0 5.5cm}, clip]{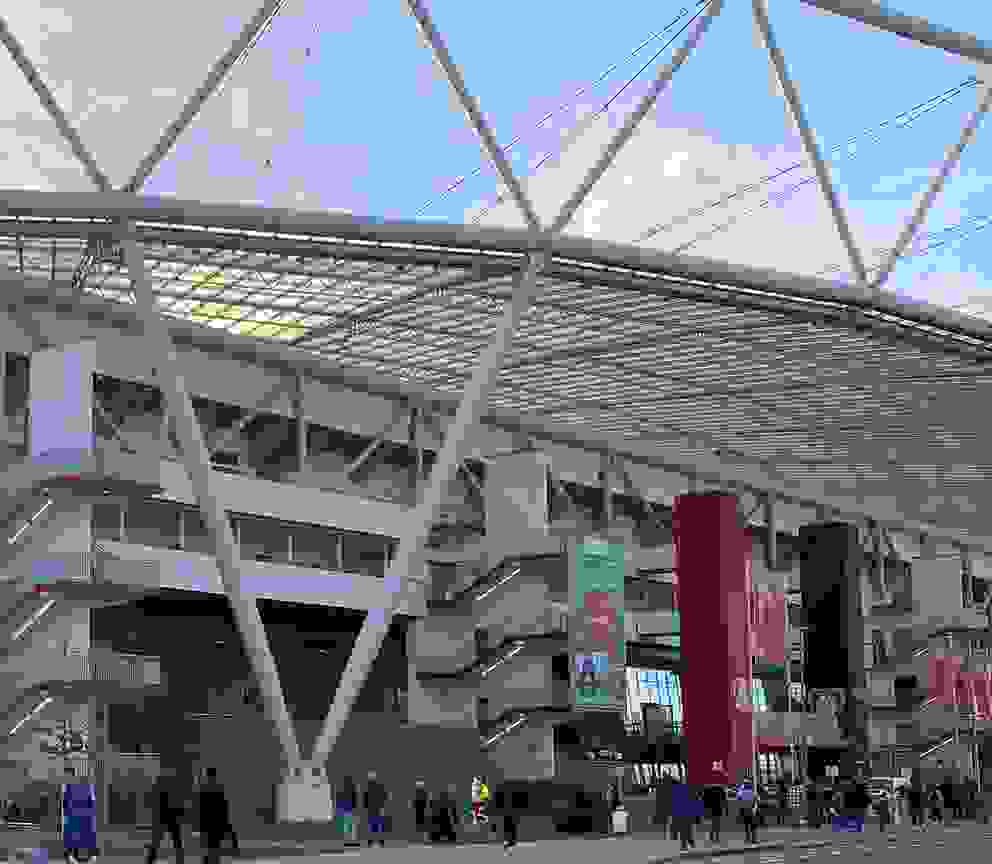}};
    \end{scope}
    \draw[thick, ->] (2.0, 0) -- (1.05, 0);
    
\end{tikzpicture}}
% \end{document}

%% file: content/method.tex
\section{Method: Differentiable JPEG Coding}
\label{subsec:jpeg}

We aim to build a continuous and differentiable approximation ($\jpeg_{\text{diff}}:[0, 255]^{3\times H\times W}\times[1, 99]\to[0, 255]^{3\times H\times W}$) of the full JPEG encoding-decoding function (\cf \cref{eq:jpeg_coding}). This approximation should accurately resemble standard (non-diff.) JPEG and yield gradients ``useful'' for gradient-based optimization. To achieve this, we first take a surrogate model approach (\cref{subsec:surrogate_approach}). Later we present the incorporation of the STE technique (\cref{subsec:ste_approach}).

We noticed that existing approaches just focus on finding ``useful'' differentiable surrogates of the rounding function used for quantization; however, other discretizations and bounds are not modeled. These operations include the clipping as well as the discretization of the quantization table, the discretization of the quantization table scaling, and the bounding of the output/coded image. We present differentiable approximations for modeling all five operations. Note for operations not described (\eg, DCT), we use their naive continuous generalization (\cf \cref{sec:jpegstandard}). Since the gradient is not affected by the lossless encoding/decoding (\cf \cref{fig:jpeg}), we follow existing approaches and neglect these steps.

\subsection{Differentiable JPEG surrogate} \label{subsec:surrogate_approach}

\paragraph{Differentiable quantization.} For approximating the quantization operation, which uses the rounding function, we utilize the polynomial approximation $\lfloor x\rceil+(x - \lfloor x\rceil)^{3}$ by Shin \etal~\cite{Shin2017}. While other approximations exist (\eg, sigmoid func.), we later show that this design choice leads to superior performance over other approximations (\cf \cref{subsec:additional_ablations}).

\paragraph{Differentiable QT scale floor.} Non-differentiable standard JPEG computes the quantization table scaling $s\!\left(\q\right)$ based on the JPEG quality with integer arithmetic (\cf \cref{eq:qt_scale}). This is equivalent to computing $s$ with float precision and applying the floor function. To model this operation in a differentiable manner, we introduce a differentiable floor approach. Note the original scaling approach (\cf \cref{eq:qt_scale}) is used, not the approach by Shin \etal~\cite{Shin2017}.

Our differentiable floor function makes use of the relation between the rounding and floor function. We can express the floor function as a shifted version of the rounding function $\lfloor x - 0.5\rceil=\lfloor x\rfloor$. Based on this property, we can use the polynomial rounding approach to approximate the floor function by $\lfloor x - 0.5\rceil+(x - 0.5 - \lfloor x - 0.5\rceil)^{3}$. We later validate this design choice against other approximations (\cf \cref{subsec:additional_ablations}).

\paragraph{Differentiable QT floor.} Since the quantization table is included in every JPEG file, the JPEG standard requires the QT to include integer values. The standard (non-diff.) JPEG implementation ensures this by using integer arithmetic. This is equivalent to applying the floor function to QT after scaling. To ensure gradient propagation, we apply the proposed floor approximation to the scale QT.

\paragraph{Differentiable QT clipping.} Based on the JPEG standard~\cite{Wallace1992}, the quantization table is bounded to the integer range $\{1, \ldots, 255\}^{8\times 8}$. Utilizing low JPEG qualities (strong compression) can lead to values outside of this range, even when utilizing the standard QT. To ensure values approximately within this range, we propose a differentiable (soft) clipping operation $\overline{\operatorname{clip}}$.
\begin{equation}\label{eq:diff_clip}
    \overline{\operatorname{clip}}\!\left(x\right)\begin{cases} 
    	x\hphantom{\gamma}\,\;\;\;\;\;\;\;\;\;\;\;\;\;\;\;\;\;\;\;\;\;\;\;\,\, \text{if}\, x\in [b_{\text{min}}, b_{\text{max}}] \\[-1.5pt]
    	b_{\text{min}} + \gamma\,(x - b_{\text{min}})\;\, \text{if}\, x<b_{\text{min}}\\[-1.5pt]
    	b_{\text{max}} + \gamma\,(x - b_{\text{max}})\; \text{if}\, x>b_{\text{max}}
    \end{cases}\!\!\!\!\!\!\!,\gamma\in(0,1].
\end{equation}
This soft approximation ensures a non-zero gradient of $x$ when outside of the range $[b_{\text{min}}, b_{\text{max}}]$. We set the scale parameter $\gamma$ to $10^{-3}$.

\paragraph{Differentiable output clipping.} Similar to the input image, the output image is bounded to the pixel range of $\{0,\ldots,255\}$. Depending on the image content and the applied JPEG quality, values outside of this range can occur. To approximately adhere to this range, we also apply the proposed differentiable clipping to the output/coded image.

\subsection{Differentiable JPEG coding with STE} \label{subsec:ste_approach}

Instead of differentiably approximating all discretizations and bounds both in the forward and backward pass, we can also take advantage of the STE technique. In our STE-based differentiable JPEG approach, we utilize the true rounding, floor, and clipping functions in the forward pass. However, instead of using a constant gradient of one, as done by standard STE, we employ the gradient of the proposed approximations during backpropagation. This approach leads to a reduced error of the forward function since the true function and not an approximation is used. We later show that our STE approach can be beneficial in certain settings. We also show that using the gradient of the proposed approximations is more effective than standard STE.

\section{Evaluation}

Evaluating a differentiable JPEG implementation comes in two different flavors. First, evaluating the performance of the forward mapping and second, the validation of gradients obtained from the differentiable JPEG approach. While evaluating the forward mapping is well-defined, validating the effectiveness of the backward function is non-trivial.

\paragraph{Forward function evaluation.} We evaluate the performance of the forward mapping by measuring the similarity between the coded image obtained by the differentiable approach and the coded image of the reference implementation. We use the SSIM~\cite{Wang2004} and the PSNR for evaluation.

\paragraph{Backward function evaluation.} We aim to showcase the ``usefulness'' of gradients obtained by the backward function. Taking inspiration from Shin \etal~\cite{Shin2017} we utilize adversarial attack experiments to showcase the quality and ``usefulness'' of gradients in the context of gradient-based optimization. Adversarial examples are crafted through the mapping $C\left(\jpeg_{\text{diff}}\left(\mathbf{I}, q\right)\right)=\mathbf{p}$, composed of an ImageNet classifier $C$ (\eg, ResNet-50~\cite{He2016}) and a differentiable JPEG function $\jpeg_{\text{diff}}$, conditioned on a given JPEG quality $q$. We aim to craft an adversarial image $\mathbf{I}_{\text{adv}}$, from the original image $\mathbf{I}$, s.t. the prediction $p\in[0, 1]^{c}$ over $c$ classes is deteriorated. Note, while crafting the adversarial example using a differentiable JPEG approach we validate the effectiveness of the adversarial example by using the non-differentiable JPEG reference implementation.

We consider two adversarial attack techniques, the fast gradient sign method (FGSM)~\cite{Goodfellow2014} and the iterative fast gradient sign method (IGSM)~\cite{Kurakin2017, Kurakin2018}. FGSM in the non-targeted setting crafts an adversarial example by $\mathbf{I}_{\text{adv}}=\mathbf{I}+\epsilon\cdot\sign(\Delta_\mathbf{I}[\mathcal{L}(y, C\left(\jpeg_{\text{diff}}\left(\mathbf{I}, q\right)\right))])$, where $\mathcal{L}$ is the cross-entropy loss and $y$ the true class label. IFGSM performs FGSM for $N$ iterations and uses $\frac{\epsilon}{N}$ as the update size. Both FGSM and IFGSM ensure that $\|\mathbf{I}_{\text{adv}} - \mathbf{I}\|_{\infty}\leq\epsilon$.

We argue that in order to generate an effective adversarial example, $\jpeg_{\text{diff}}$ needs to produce ``useful'' gradients. The more effective the adversarial images are in deteriorating the prediction (measured by accuracy), the more ``useful'' the gradients of $\jpeg_{\text{diff}}$ become. Note, since $C$ consumes the JPEG-coded image the resulting gradient is also partly dependent on the forward performance of $\jpeg_{\text{diff}}$.

\paragraph{Vanishing gradient evaluation.} While it is not possible to directly measure the ``usefulness'' of gradients, we can evaluate the gradients' ability to adhere to desired properties in gradient-based optimization. For local and global minima, it is desirable that gradients vanish. While the exact position of local and global minima \wrt to a differentiable JPEG approach is not known, we can measure the gradient magnitude at positions that are desired to be local or global minima. Optimally, the gradient at these positions vanishes. In particular, we compute the L1 loss between the differentiable and the reference JPEG $\mathcal{L}_{1}\!\left(\jpeg_{\text{diff}}\left(\mathbf{I}, \q\right), \jpeg_{\text{r}}\left(\mathbf{I},\q\right)\right)$. Subsequently, estimate the gradient norms \wrt to both the JPEG quality $\|\Delta_{\q}\mathcal{L}_{1}\|$ and the (internal) standard QT $\|\Delta_{\qt_{Y, C}}\mathcal{L}_{1}\|$ (luma and chroma table). We average the gradient norms over both the different integer JPEG quality ranges and the dataset $\mathcal{D}$.

%% file: content/experiments.tex
\section{Experiments}
\label{sec:experiments}

\paragraph{Datasets.} For our adversarial attack experiments, we utilize $5\si{\kilo\relax}$ randomly chosen images of the ImageNet-1k validation set (ILSVRC 2012)~\cite{Russakovsky2015}. For all other experiments, we use the Set14 dataset~\cite{Zeyde2012}, composed of 14 RGB images ranging from natural to document-like images.

\paragraph{Implementation details.} For all experiments, we utilize the OpenCV~\cite{Bradski2000} JPEG implementation as a reference. We utilize Kornia~\cite{Riba2020} for computing the SSIM. The utilized SSIM patch size is 11. Adversarial attack experiments are conducted using a ResNet-50 from torchvision~\cite{Torchvision2016} supervised trained on ImageNet-1k. For IFGSM experiments, we utilize $N=10$ iterations. The step parameter $\epsilon$ is varied between experiments. After each attack iteration, we hard clip the image to the valid pixel range of $[0, 255]$.

\paragraph{Baselines.} We compare against the surrogate-based approaches by Xing \etal~\cite{Xing2021} and Shin \etal~\cite{Shin2017}. We also compare against the STE-based approach of Xie \etal~\cite{Xie2022}. Since Xie \etal are not modeling the JPEG quality, we extend the approach with the JPEG quality mapping of Xing \etal~\cite{Xing2021}. Both the approach of Shin \etal and Xie \etal offer no code, we have reimplemented both in PyTorch~\cite{Paszke2019}. For the approach of Xing \etal~\cite{Xing2021}, we use the official code. We noticed a bug in the official code (wrong $\qt$ transposition). We run experiments with the debug code. Note due to the very limited control (JPEG quality can not be set) and stochasticity, we do not consider noise-based approaches.

\subsection{Forward function results}

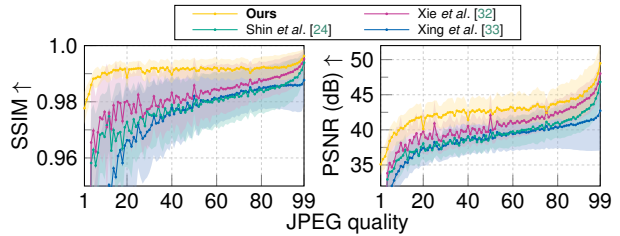
\begin{figure}[ht!]
    \centering
    \input{artwork/forward_performance_small}
    \caption{\textbf{Forward function performance.} Performance of approximating the reference JPEG implementation (OpenCV~\cite{Bradski2000}) for different JPEG qualities. Mean \& one standard deviation shown.}
    \label{fig:forward_performance}
    \vspace{-0.575em}
\end{figure}

We evaluate the ability to resemble the reference JPEG implementation of existing approaches against our differentiable JPEG approach (w/ STE). Our approach outperforms existing approaches over the whole JPEG quality range (\cf \cref{fig:forward_performance}). The performance gap between differentiable methods becomes smaller for high JPEG quality values, but still, our approach leads to superior performance.

\begin{figure}[ht!]
    \centering
    \input{artwork/forward_performance_low_cs_small}
    \caption{\textbf{Forward function performance for strong compression.} Performance of approximating the reference JPEG implementation (OpenCV~\cite{Bradski2000}) for low JPEG qualities. Mean \& one standard deviation shown.}
    \label{fig:forward_performance_low_cs}
    \vspace{-0.575em}
\end{figure}
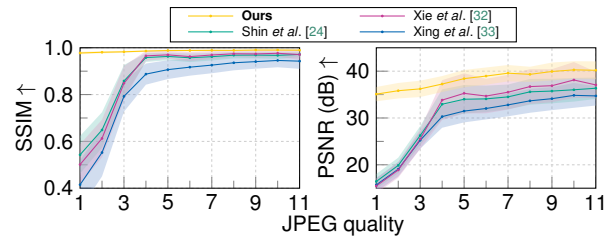

For small JPEG qualities, we observe a significant disparity in performance between methods. While our approach still approximates standard JPEG well, existing approaches fail and produce coded images vastly different from the reference implementation. This is showcased in \cref{fig:first} and quantitatively analyzed in \cref{fig:forward_performance_low_cs} and \cref{tab:ablation_forward}.

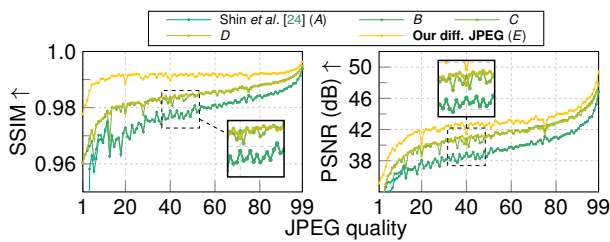
\begin{figure}[t]
    \centering
    \input{artwork/ablation_small}
    \caption{\textbf{Ablation study graph.} JPEG quality-wise forward function performance for different configurations (\cf \cref{tab:ablation_forward}).}
    \label{fig:ablation_forward}
    \vspace{-0.575em}
\end{figure}

\paragraph{Forward function ablation results.} To understand what makes our differentiable JPEG implementation effective in resembling the reference implementation, we conduct an ablation study (\cf \cref{tab:ablation_forward} and \cref{fig:ablation_forward}). We gradually add our introduced components (config. \emph{A} to \emph{F}). All our novel components improve performance while our full configuration (config. \emph{E}) performs best among non-STE approaches. Using STE (config. \emph{F}) further improves forward performance, leading to coded images perceptually indistinguishable from the reference implementation.

\begin{table}[t!]
    \centering
    \scriptsize
    \setlength\tabcolsep{0.25pt}
    \input{table/ablation}
    \caption{\textbf{Forward function performance summary \& ablation.} To ablate our approach, we gradually add our novel components to Shin \etal~\cite{Shin2017}. We also report the performance of other diff. approaches. STE-based approaches marked in \colorindicator{gray}{tud0a}.}
    \label{tab:ablation_forward}
    \vspace{-0.575em}
\end{table}

\cref{fig:ablation_forward} showcases the effect of all introduced components for each (integer) JPEG quality. We observe that especially differentially clipping the coded output image (config. \emph{E}) improves performance and, in particular, leads to strong results for small JPEG qualities.

Using STE (config. \emph{F}) leads to a superior forward function performance over our non-STE approach (config. \emph{E}, \cf \cref{fig:ablation_forward}). As showcased in \cref{fig:forward_performance_ste}, our diff. STE JPEG approach especially outperforms our differentiable JPEG approach w/o STE for low JPEG qualities.

\begin{figure}[ht!]
    \centering
    \input{artwork/forward_performance_ste_small}
    \caption{\textbf{Forward function performance with STE.} Performance of approximating the reference JPEG implementation (OpenCV~\cite{Bradski2000}) for different JPEG qualties. Mean \& one SD shown.}
    \label{fig:forward_performance_ste}
    \vspace{-0.575em}
\end{figure}

\subsection{Backward function results}

\paragraph{Adversarial attack results.} We craft adversarial examples to showcase the ``usefulness'' of gradients derived by differentiable JPEG approaches. When using FGSM with $\epsilon=3$, our differentiable JPEG w/o STE consistently leads to superior results over our approach w/ STE and other differentiable approaches (\cf \cref{tab:fgsm} (top)). When increasing $\epsilon$ to $9$, our differentiable JPEG w/ STE scores slightly better than our approach w/o STE (\cf \cref{tab:fgsm} (bottom)). 

\begin{table}[ht!]
    \centering
    \scriptsize
    \setlength\tabcolsep{0.25pt}
    \input{table/fgsm}
    \caption{\textbf{FGSM attack results summary.} Summarized top-1 and top-5 accuracy after FGSM attack for different JPEG quality ranges. We report results for both $\epsilon=3$ and $\epsilon=9$. As a reference, we also report accuracies for no attack performed.}
    \label{tab:fgsm}
    \vspace{-0.575em}
\end{table}

When using IFGSM to craft adversarial examples, the results are consistently in favor of our differentiable JPEG approach w/o STE (\cf \cref{tab:ifgsm}). Interestingly, the approach by Xing \etal leads to predominately poor adversarial examples. We explain this result by the use of the Fourier rounding approximation which is highly non-monotonic.

\begin{table}[ht!]
    \centering
    \scriptsize
    \setlength\tabcolsep{0.25pt}
    \input{table/ifgsm}
    \caption{\textbf{IFGSM attack results summary.} Summarized top-1 and top-5 accuracy results after IFGSM attack for multiple JPEG quality ranges and different differentiable JPEG approaches. We report accuracy results for both $\epsilon=3$ and $\epsilon=9$.}
    \label{tab:ifgsm}
    \vspace{-0.575em}
\end{table}

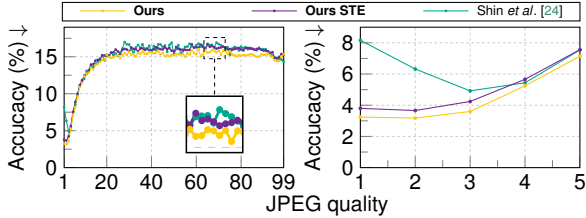
\begin{figure}[t!]
    \centering
    \input{artwork/ifgsm_small}
    \caption{\textbf{IFGSM attack results.} Top-1 accuracy after IFGSM attack with our approach (w/ \& w/o STE) \vs Shin \etal~\cite{Shin2017}.}
    \label{fig:ifgsm}
    \vspace{-0.575em}
\end{figure}

Our approach w/o STE leads to particularly strong attack results for low JPEG qualities compared to other approaches (\cf \cref{tab:ifgsm}). Our approaches (w/ \& w/o STE) lead to strong adversarial images for a JPEG quality of $1$, $2$, and $3$, while Shin \etal suffer to produce effective adversarial images (\cf \cref{fig:ifgsm}). In general, our approach w/ STE leads to a stronger forward performance, while a better backward performance is achieved w/o STE.

\paragraph{Vanishing gradient experiment.} Both our differentiable JPEG w/ and w/o STE lead to lower gradient norms at the desired minima (\cf \cref{tab:vanishing_gradient}). The approach by Xing \etal leads to particularly large gradient norms. We suspect again the choice of rounding function approximation (Fourier-based) to be a major contributor to these results.

\begin{table}[ht!]
    \centering
    \scriptsize
    \setlength\tabcolsep{0.25pt}
    \input{table/vanishing_gradient}
    \caption{\textbf{Vanishing gradient results.} Average gradient norms at desired global minima. Gradient norms are averaged over the integer JPEG quality range $Q$ and the Set14~\cite{Zeyde2012} dataset $\mathcal{D}$.}
    \label{tab:vanishing_gradient}
    \vspace{-0.575em}
\end{table}

\subsection{Additional ablation results} \label{subsec:additional_ablations}

\paragraph{Which rounding/floor approximation to use?} In \cref{tab:round_function_forward}, we analyze the effect of different differentiable rounding and floor function approximations on the forward performance. Both the linear and the polynomial differentiable approximation perform best, with a slight advantage for the polynomial approach.

\begin{table}[ht!]
    \centering
    \scriptsize
    \setlength\tabcolsep{0.25pt}
    \input{table/round_function_forward_function}
    \caption{\textbf{Rounding/floor ablation (forward function).} Performance of approximating the reference OpenCV~\cite{Bradski2000} JPEG. For all experiments, our diff. JPEG w/o STE was used; only the differentiable rounding/floor approx. is varied. Eq. describe the rounding approximations; for the floor approx., we shift $x$ by $-0.5$.}
    \label{tab:round_function_forward}
    \vspace{-0.575em}
\end{table}

As in terms of backward function performance, the polynomial approximation also leads to the best adversarial examples (\cf \cref{tab:round_function_ifgsm}). This suggests gradients derived from the polynomial approximation are more ``useful'' than from other approximations. While leading to a fair forward performance (\cf \cref{tab:round_function_forward}), the Fourier approximation leads to poor attack results, thus also yields worse gradients. This result aligns with the attack results by Xing \etal (\cf \cref{tab:fgsm} \& \cref{tab:ifgsm}) also employing the Fourier approximation.

\begin{table}[ht!]
    \centering
    \scriptsize
    \setlength\tabcolsep{0.25pt}
    \input{table/round_function_ifgsm}
    \caption{\textbf{Rounding/floor ablation (IFGSM).} IFGSM attack results for various differentiable rounding/floor approximations. IFGSM w/ $\epsilon=3$ used. For all experiments; our diff. JPEG was used, only the rounding/floor approximation was varied.}
    \label{tab:round_function_ifgsm}
    \vspace{-0.575em}
\end{table}

When considering both forward and backward performance (\cf \cref{tab:round_function_forward} \& \cref{tab:round_function_ifgsm}), the best choice for approximating the rounding and floor function is the polynomial approach. These results might be explained by the fact the polynomial approximation strikes a vital trade-off between approximation error (\wrt the rounding/floor function), monotonicity, and gradient magnitude.

\paragraph{Which STE backward approximation to use?} Standard STE assumes a constant gradient in the backward pass, while our STE approach uses a closer approximation of the true derivative. When using our proposed surrogate-based STE, we observe substantial gains in adversarial attack performance over standard STE (\cf \cref{tab:ste_backward_function}), indicating that our approach leads to more ``useful'' gradients.

\begin{table}[ht!]
    \centering
    \scriptsize
    \setlength\tabcolsep{0.25pt}
    \input{table/ste_backward_function}
    \caption{\textbf{STE backward ablation (IFGSM).} IFGSM attack results for different STE backward approaches. IFGSM w/ $\epsilon=3$ used. For all experiments, our differentiable JPEG w/ STE was used; only the STE backward approach was varied.}
    \label{tab:ste_backward_function}
    \vspace{-0.575em}
\end{table}

%% file: artwork/forward_performance_small.tex
\begin{tikzpicture}[every node/.style={font=\footnotesize}]
    \node[anchor=center] at (0.42\columnwidth, -0.5) {JPEG quality};
	\begin{groupplot}[
        group style={
            group name=forwardperf,
            group size=2 by 1,
            ylabels at=edge left,
            horizontal sep=29pt,
            vertical sep=8pt,
        },
        legend style={nodes={scale=0.5}},
        height=3.45cm,
        width=0.54\columnwidth,
        grid=both,
        xtick pos=bottom,
        ytick pos=left,
        grid style={line width=.1pt, draw=white, dash pattern=on 1pt off 1pt},
        major grid style={line width=.15pt,draw=gray!40, dash pattern=on 1pt off 1pt},
        minor tick num=1,
        xmin=1,
        xticklabel shift={-1pt},
        yticklabel shift={-1pt},
        xmax=99,
        ylabel shift=-4.5pt,
        xtick={
            1, 20, 40, 60, 80, 99
        },
        xticklabels={
            1, 20, 40, 60, 80, 99
        },
        ticklabel style = {font=\footnotesize},
        ]
        % SSIM curve
        \nextgroupplot[
        ylabel=SSIM $\uparrow$,
        ytick={0.96, 0.98, 1.0},
        yticklabels={0.96, 0.98, 1.0},
        ymin=0.95,   
        ymax=1,
        ylabel style={at={(-0.20,0.5)}},
        ]
        
        \addplot[color=tud1b, mark=*, mark size=0.3pt] table[x=cs, y=ssim] {xing_forward.dat};
        \addplot[name path=upper,draw=none] table[x=cs, y expr=\thisrow{ssim}+1*\thisrow{ssimstd}] {xing_forward.dat};
        \addplot[name path=lower,draw=none] table[x=cs, y expr=\thisrow{ssim}-1*\thisrow{ssimstd}] {xing_forward.dat};
        \addplot[fill=tud1b!50, fill opacity=0.35] fill between[of=upper and lower];
        
        \addplot[color=tud3b, mark=*, mark size=0.3pt] table[x=cs, y=ssim] {shin_forward.dat};
        \addplot[name path=upper,draw=none] table[x=cs, y expr=\thisrow{ssim}+1*\thisrow{ssimstd}] {shin_forward.dat};
        \addplot[name path=lower,draw=none] table[x=cs, y expr=\thisrow{ssim}-1*\thisrow{ssimstd}] {shin_forward.dat};
        \addplot[fill=tud3b!50, fill opacity=0.35] fill between[of=upper and lower];
        
        \addplot[color=tud10a, mark=*, mark size=0.3pt] table[x=cs, y=ssim] {xie_forward.dat};
        \addplot[name path=upper,draw=none] table[x=cs, y expr=\thisrow{ssim}+1*\thisrow{ssimstd}] {xie_forward.dat};
        \addplot[name path=lower,draw=none] table[x=cs, y expr=\thisrow{ssim}-1*\thisrow{ssimstd}] {xie_forward.dat};
        \addplot[fill=tud10a!50, fill opacity=0.35] fill between[of=upper and lower];
        
        \addplot[color=tud6b, mark=*, mark size=0.3pt] table[x=cs, y=ssim] {ours_forward.dat};
        \addplot[name path=upper,draw=none] table[x=cs, y expr=\thisrow{ssim}+1*\thisrow{ssimstd}] {ours_forward.dat};
        \addplot[name path=lower,draw=none] table[x=cs, y expr=\thisrow{ssim}-1*\thisrow{ssimstd}] {ours_forward.dat};
        \addplot[fill=tud6b!50, fill opacity=0.35] fill between[of=upper and lower];
        
        % PSNR curve
        \nextgroupplot[
        ylabel=PSNR (dB) $\uparrow$,
        ytick={35, 40, 45, 50},
        yticklabels={35, 40, 45, 50},
        ymin=32,   
        ymax=52,
        ylabel style={at={(-0.13,0.5)}},
        ]
        
        \addplot[color=tud1b, mark=*, mark size=0.3pt] table[x=cs, y=psnr] {xing_forward.dat};\label{pgfplots:xing_forward};
        \addplot[name path=upper,draw=none] table[x=cs, y expr=\thisrow{psnr}+1*\thisrow{psnrstd}] {xing_forward.dat};
        \addplot[name path=lower,draw=none] table[x=cs, y expr=\thisrow{psnr}-1*\thisrow{psnrstd}] {xing_forward.dat};
        \addplot[fill=tud1b!50, fill opacity=0.35] fill between[of=upper and lower];
        
        \addplot[color=tud3b, mark=*, mark size=0.3pt] table[x=cs, y=psnr] {shin_forward.dat};\label{pgfplots:shin_forward};
        \addplot[name path=upper,draw=none] table[x=cs, y expr=\thisrow{psnr}+1*\thisrow{psnrstd}] {shin_forward.dat};
        \addplot[name path=lower,draw=none] table[x=cs, y expr=\thisrow{psnr}-1*\thisrow{psnrstd}] {shin_forward.dat};
        \addplot[fill=tud3b!50, fill opacity=0.35] fill between[of=upper and lower];
        
        \addplot[color=tud10a, mark=*, mark size=0.3pt] table[x=cs, y=psnr] {xie_forward.dat};\label{pgfplots:xie_forward};
        \addplot[name path=upper,draw=none] table[x=cs, y expr=\thisrow{psnr}+1*\thisrow{psnrstd}] {xie_forward.dat};
        \addplot[name path=lower,draw=none] table[x=cs, y expr=\thisrow{psnr}-1*\thisrow{psnrstd}] {xie_forward.dat};
        \addplot[fill=tud10a!50, fill opacity=0.35] fill between[of=upper and lower];
        
        \addplot[color=tud6b, mark=*, mark size=0.3pt] table[x=cs, y=psnr] {ours_forward.dat};\label{pgfplots:ours_forward};
        \addplot[name path=upper,draw=none] table[x=cs, y expr=\thisrow{psnr}+1*\thisrow{psnrstd}] {ours_forward.dat};
        \addplot[name path=lower,draw=none] table[x=cs, y expr=\thisrow{psnr}-1*\thisrow{psnrstd}] {ours_forward.dat};
        \addplot[fill=tud6b!50, fill opacity=0.35] fill between[of=upper and lower];
        
	\end{groupplot}

    \node[draw,fill=white, inner sep=0.5pt, anchor=center] at (0.42\columnwidth, 2.175) {\tiny
    \begin{tabular}{
    p{0.3cm}p{\widthof{Shin \etal \cite{Shin2017}}}
    p{0.3cm}p{\widthof{Xing \etal \cite{Xing2021}}}}
    \ref*{pgfplots:ours_forward} & \textbf{Ours} & \ref*{pgfplots:xie_forward} & Xie \etal~\cite{Xie2022} \\
    \ref*{pgfplots:shin_forward} & Shin \etal \cite{Shin2017} & \ref*{pgfplots:xing_forward} & Xing \etal \cite{Xing2021}\\
    \end{tabular}};
\end{tikzpicture}

%% file: artwork/forward_performance_low_cs_small.tex
\begin{tikzpicture}[every node/.style={font=\footnotesize}]
    \node[anchor=center] at (0.42\columnwidth, -0.5) {JPEG quality};
	\begin{groupplot}[
        group style={
            group name=forwardperflow,
            group size=2 by 2,
            ylabels at=edge left,
            horizontal sep=29pt,
            vertical sep=8pt,
        },
        enlargelimits=false,
        legend style={nodes={scale=0.5}},
        height=3.45cm,
        width=0.54\columnwidth,
        grid=both,
        xtick pos=bottom,
        xticklabel shift={-1pt},
        yticklabel shift={-1pt},
        ytick pos=left,
        grid style={line width=.1pt, draw=white, dash pattern=on 1pt off 1pt},
        major grid style={line width=.15pt,draw=gray!40, dash pattern=on 1pt off 1pt},
        minor tick num=1,
        xmin=1,
        xmax=11,
        ylabel shift=-4.5pt,
        xtick={
            1, 3, 5, 7, 9, 11
        },
        xticklabels={
            1, 3, 5, 7, 9, 11
        },
        ticklabel style = {font=\footnotesize},
        ]
        % SSIM curve
        \nextgroupplot[
        ylabel=SSIM $\uparrow$,
        ytick={0.4, 0.6, 0.8, 1.0},
        yticklabels={0.4, 0.6, 0.8, 1.0},
        ymin=0.4,   
        ymax=1,
        ylabel style={at={(-0.16,0.5)}},
        ]
        
        \addplot[color=tud1b, mark=*, mark size=0.3pt] table[x=cs, y=ssim] {xing_forward.dat};
        \addplot[name path=upper,draw=none] table[x=cs, y expr=\thisrow{ssim}+1*\thisrow{ssimstd}] {xing_forward.dat};
        \addplot[name path=lower,draw=none] table[x=cs, y expr=\thisrow{ssim}-1*\thisrow{ssimstd}] {xing_forward.dat};
        \addplot[fill=tud1b!50, fill opacity=0.35] fill between[of=upper and lower];
        
        \addplot[color=tud3b, mark=*, mark size=0.3pt] table[x=cs, y=ssim] {shin_forward.dat};
        \addplot[name path=upper,draw=none] table[x=cs, y expr=\thisrow{ssim}+1*\thisrow{ssimstd}] {shin_forward.dat};
        \addplot[name path=lower,draw=none] table[x=cs, y expr=\thisrow{ssim}-1*\thisrow{ssimstd}] {shin_forward.dat};
        \addplot[fill=tud3b!50, fill opacity=0.35] fill between[of=upper and lower];
        
        \addplot[color=tud10a, mark=*, mark size=0.3pt] table[x=cs, y=ssim] {xie_forward.dat};
        \addplot[name path=upper,draw=none] table[x=cs, y expr=\thisrow{ssim}+1*\thisrow{ssimstd}] {xie_forward.dat};
        \addplot[name path=lower,draw=none] table[x=cs, y expr=\thisrow{ssim}-1*\thisrow{ssimstd}] {xie_forward.dat};
        \addplot[fill=tud10a!50, fill opacity=0.35] fill between[of=upper and lower];
        
        \addplot[color=tud6b, mark=*, mark size=0.3pt] table[x=cs, y=ssim] {ours_forward.dat};
        \addplot[name path=upper,draw=none] table[x=cs, y expr=\thisrow{ssim}+1*\thisrow{ssimstd}] {ours_forward.dat};
        \addplot[name path=lower,draw=none] table[x=cs, y expr=\thisrow{ssim}-1*\thisrow{ssimstd}] {ours_forward.dat};
        \addplot[fill=tud6b!50, fill opacity=0.35] fill between[of=upper and lower];
        
        % PSNR curve
        \nextgroupplot[
        ylabel=PSNR (dB) $\uparrow$,
        ytick={20, 30, 40},
        yticklabels={20, 30, 40},
        ymin=15,   
        ymax=45,
        ylabel style={at={(-0.13,0.5)}},
        ]
        
        \addplot[color=tud1b, mark=*, mark size=0.3pt] table[x=cs, y=psnr] {xing_forward.dat}; \label{pgfplots:xing_forward_low_cs};
        \addplot[name path=upper,draw=none] table[x=cs, y expr=\thisrow{psnr}+1*\thisrow{psnrstd}] {xing_forward.dat};
        \addplot[name path=lower,draw=none] table[x=cs, y expr=\thisrow{psnr}-1*\thisrow{psnrstd}] {xing_forward.dat};
        \addplot[fill=tud1b!50, fill opacity=0.35] fill between[of=upper and lower];
        
        \addplot[color=tud3b, mark=*, mark size=0.3pt] table[x=cs, y=psnr] {shin_forward.dat}; \label{pgfplots:shin_forward_low_cs};
        \addplot[name path=upper,draw=none] table[x=cs, y expr=\thisrow{psnr}+1*\thisrow{psnrstd}] {shin_forward.dat};
        \addplot[name path=lower,draw=none] table[x=cs, y expr=\thisrow{psnr}-1*\thisrow{psnrstd}] {shin_forward.dat};
        \addplot[fill=tud3b!50, fill opacity=0.35] fill between[of=upper and lower];
        
        \addplot[color=tud10a, mark=*, mark size=0.3pt] table[x=cs, y=psnr] {xie_forward.dat}; \label{pgfplots:xie_forward_low_cs};
        \addplot[name path=upper,draw=none] table[x=cs, y expr=\thisrow{psnr}+1*\thisrow{psnrstd}] {xie_forward.dat};
        \addplot[name path=lower,draw=none] table[x=cs, y expr=\thisrow{psnr}-1*\thisrow{psnrstd}] {xie_forward.dat};
        \addplot[fill=tud10a!50, fill opacity=0.35] fill between[of=upper and lower];
        
        \addplot[color=tud6b, mark=*, mark size=0.3pt] table[x=cs, y=psnr] {ours_forward.dat}; \label{pgfplots:ours_forward_low_cs};
        \addplot[name path=upper,draw=none] table[x=cs, y expr=\thisrow{psnr}+1*\thisrow{psnrstd}] {ours_forward.dat};
        \addplot[name path=lower,draw=none] table[x=cs, y expr=\thisrow{psnr}-1*\thisrow{psnrstd}] {ours_forward.dat};
        \addplot[fill=tud6b!50, fill opacity=0.35] fill between[of=upper and lower];
        
	\end{groupplot}
 
    \node[draw,fill=white, inner sep=0.5pt, anchor=center] at (0.42\columnwidth, 2.175) {\tiny
    \begin{tabular}{
    p{0.3cm}p{\widthof{Shin \etal \cite{Shin2017}}}
    p{0.3cm}p{\widthof{Xing \etal \cite{Xing2021}}}}
    \ref*{pgfplots:ours_forward_low_cs} & \textbf{Ours} & \ref*{pgfplots:xie_forward_low_cs} & Xie \etal~\cite{Xie2022} \\
    \ref*{pgfplots:shin_forward_low_cs} & Shin \etal \cite{Shin2017} & \ref*{pgfplots:xing_forward_low_cs} & Xing \etal \cite{Xing2021}\\
    \end{tabular}};
\end{tikzpicture}

%% file: table/ablation.tex
% \begin{tabular*}{\columnwidth}{@{\extracolsep{\fill}}l@{\hskip 0.25em}S[table-format=1.3]S[table-format=1.3]S[table-format=1.3]S[table-format=2.2]S[table-format=2.2]S[table-format=2.2]@{}}
% 	\toprule
% 	& \multicolumn{3}{c}{SSIM $\uparrow$} & \multicolumn{3}{c}{PSNR $\uparrow$}  \\
%     \cmidrule{2-4}\cmidrule{5-7}
% 	{Configuration\hspace{0.8cm}$\q$ range $\to$\hspace{-0.4cm}} & {1-99} & {1-10} & {11-99} & {1-99} & {1-10} & {10-99} \\
%     \midrule
%     {\footnotesize\emph{\hphantom{A}}}$\;\;$Xing \etal~\cite{Xing2021} & 0.961 & 0.833 & 0.977 & 38.10 & 29.45 & 39.19 \\
% 	\midrule
% 	{\scriptsize\emph{A}}$\;\;$Shin \etal~\cite{Shin2017} & 0.974 & 0.892 & 0.984 & 40.26 & 31.78 & 41.32 \\
%     {\scriptsize\emph{B}}$\;\;$+ diff. linear quantization & 0.967 & 0.883 & 0.978 & 38.45 & 30.88 & 39.39 \\
%     {\scriptsize\emph{C}}$\;\;$+ diff. linear round YCbCr & 0.968 & 0.884 & 0.979 & 38.60 & 31.01 & 39.55 \\
%     {\scriptsize\emph{D}}$\;\;$+ diff. QT clipping & 0.978 & 0.968 & 0.979 & 39.12 & 35.59 & 39.56 \\
%     {\scriptsize\emph{E}}$\;\;$+ diff. QT linear floor & 0.982 & 0.974 & 0.985 & 41.36 & 36.94 & 41.92 \\
%     {\scriptsize\emph{F}}$\;\;$+ diff. QT scale linear floor & 0.984 & 0.974 & 0.986 & 41.56 & 37.01 & 42.13 \\
%     \midrule
%     {\scriptsize\emph{G}}$\;\;$+ diff. output clipping \\ (our differentiable JPEG) & 0.991 & 0.987 & 0.992 & 42.60 & 38.28 & 43.14 \\
%     \midrule
%     {\scriptsize\emph{H}}$\;\;$+ straight-through estimator \\ (our differentiable STE JPEG) & 0.993 & 0.993 & 0.992 & 43.49 & 41.14 & 43.78 \\
% 	\bottomrule
% \end{tabular*}
\renewcommand\arraystretch{0.98}
\begin{tabular*}{\columnwidth}{@{\extracolsep{\fill}}ll@{\hskip 0.1em}S[table-format=1.3]S[table-format=1.3]S[table-format=1.3]S[table-format=2.2]S[table-format=2.2]S[table-format=2.2]@{}}
	\toprule
	& & \multicolumn{3}{c}{SSIM $\uparrow$} & \multicolumn{3}{c}{PSNR $\uparrow$}  \\
    \cmidrule{3-5}\cmidrule{6-8}
	\multicolumn{2}{l}{Configuration\hspace{0.78cm}$\q$ range $\to$\hspace{-0.5cm}} & {1-99} & {1-10} & {11-99} & {1-99} & {1-10} & {11-99} \\
    \midrule
    \cellcolor{tud1b!50} & \cellcolor{white} Xing \etal~\cite{Xing2021} & 0.961 & 0.833 & 0.977 & 38.10 & 29.45 & 39.19 \\
    \cellcolor{tud10a!50} & \cellcolor{tud0a} Xie \etal~\cite{Xie2022} & 0.972 & 0.884 & 0.983 & 40.02 & 31.63 & 41.07 \\
	\cellcolor{tud3b!50} {\scriptsize\emph{A}}$\,$ & \cellcolor{white} Shin \etal~\cite{Shin2017} & 0.969 & 0.888 & 0.979 & 38.71 & 31.07 & 39.66 \\
	\midrule
    % \cellcolor{tud3b6b1!50} {\scriptsize\emph{B}} & + std. QT scale (\cf \cref{eq:qt_scale}) & 0.977 & 0.905 & 0.986 & 39.68 & 32.18 & 40.61 \\
    \cellcolor{tud3b6b1!50} {\scriptsize\emph{B}} & \cellcolor{white} + diff. QT clipping & 0.978 & 0.966 & 0.979 & 39.16 & 35.10 & 39.67 \\
    \cellcolor{tud3b6b2!50} {\scriptsize\emph{C}} & \cellcolor{white} + diff. QT floor & 0.983 & 0.971 & 0.985 & 41.03 & 35.95 & 41.66 \\
    \cellcolor{tud3b6b3!50} {\scriptsize\emph{D}} & \cellcolor{white} + diff. QT scale floor & 0.984 & 0.971 & 0.986 & 41.08 & 35.96 & 41.72 \\
    \midrule
    \cellcolor{tud6b!50} {\scriptsize\emph{E}} & \cellcolor{white} + diff. output clipping & {\multirow{2}{*}{\textit{0.991}}} & {\multirow{2}{*}{\textit{0.987}}} & {\multirow{2}{*}{\textit{0.992}}} & {\multirow{2}{*}{\textit{42.60}}} & {\multirow{2}{*}{\textit{38.28}}} & {\multirow{2}{*}{\textit{43.14}}} \\
    \multicolumn{2}{l}{\textbf{(our differentiable JPEG)}} & \\
    \midrule
    \cellcolor{tud11b!50} {\scriptsize\emph{F}}& \cellcolor{tud0a} + STE (\cf \cref{subsec:ste_approach}) & {\multirow{2}{*}{\textbf{0.993}}} & {\multirow{2}{*}{\textbf{0.993}}} & {\multirow{2}{*}{\textbf{0.992}}} & {\multirow{2}{*}{\textbf{43.49}}} & {\multirow{2}{*}{\textbf{41.14}}} & {\multirow{2}{*}{\textbf{43.78}}} \\
    \multicolumn{2}{l}{\textbf{(our differentiable STE JPEG)}}\\
	\bottomrule
\end{tabular*}

%% file: artwork/ablation_small.tex
\begin{tikzpicture}[every node/.style={font=\footnotesize}, spy using outlines={black, line width=0.80mm, dashed, dash pattern=on 1.5pt off 1.5pt, magnification=1.5, size=0.75cm, connect spies, fill=white}]
    \node[anchor=center] at (0.42\columnwidth, -0.5) {JPEG quality};
	\begin{groupplot}[
        group style={
            group name=csvsdistortion,
            group size=2 by 2,
            ylabels at=edge left,
            horizontal sep=29pt,
            vertical sep=8pt,
        },
        enlargelimits=false,
        enlargelimits=false,
        legend style={nodes={scale=0.5}},
        height=3.45cm,
        width=0.54\columnwidth,
        grid=both,
        xtick pos=bottom,
        xticklabel shift={-1pt},
        yticklabel shift={-1pt},
        ytick pos=left,
        grid style={line width=.1pt, draw=white, dash pattern=on 1pt off 1pt},
        major grid style={line width=.15pt,draw=gray!40, dash pattern=on 1pt off 1pt},
        minor tick num=1,
        xmin=1,
        xmax=99,
        ylabel shift=-4.5pt,
        xtick={
            1, 20, 40, 60, 80, 99
        },
        xticklabels={
            1, 20, 40, 60, 80, 99
        },
        ticklabel style = {font=\footnotesize}
        ]
        % SSIM curve
        \nextgroupplot[
        ylabel=SSIM $\uparrow$,
        ytick={0.96, 0.98, 1.0},
        yticklabels={0.96, 0.98, 1.0},
        ymin=0.95,   
        ymax=1,
        ylabel style={at={(-0.20,0.5)}},
        ]
        \addplot[color=tud3b, mark=*, mark size=0.3pt] table[x=cs, y=ssim] {shin_forward.dat};

        \addplot[color=tud3b6b1, mark=*, mark size=0.3pt] table[x=cs, y=ssim] {ours_forward_b.dat};

        \addplot[color=tud3b6b2, mark=*, mark size=0.3pt] table[x=cs, y=ssim] {ours_forward_c.dat};

        \addplot[color=tud3b6b3, mark=*, mark size=0.3pt] table[x=cs, y=ssim] {ours_forward_d.dat};
        
        \addplot[color=tud6b, mark=*, mark size=0.3pt] table[x=cs, y=ssim] {ours_forward.dat};
        
        % PSNR curve
        \nextgroupplot[
        ylabel=PSNR (dB) $\uparrow$,
        ytick={38, 42, 46, 50},
        yticklabels={38, 42, 46, 50},
        ymin=34,   
        ymax=52,
        ylabel style={at={(-0.13,0.5)}},
        ]
        \addplot[color=tud3b, mark=*, mark size=0.3pt] table[x=cs, y=psnr] {shin_forward.dat};\label{pgfplots:shin_ablation};

        \addplot[color=tud3b6b1, mark=*, mark size=0.3pt] table[x=cs, y=psnr] {ours_forward_b.dat};\label{pgfplots:b_ablation};

        \addplot[color=tud3b6b2, mark=*, mark size=0.3pt] table[x=cs, y=psnr] {ours_forward_c.dat};\label{pgfplots:c_ablation};

        \addplot[color=tud3b6b3, mark=*, mark size=0.3pt] table[x=cs, y=psnr] {ours_forward_d.dat};\label{pgfplots:d_ablation};
        
        \addplot[color=tud6b, mark=*, mark size=0.3pt] table[x=cs, y=psnr] {ours_forward.dat};\label{pgfplots:g_ablation};
        
	\end{groupplot}

    \node[draw,fill=white, inner sep=0.5pt, anchor=center] at (0.42\columnwidth, 2.175) {\tiny
    \begin{tabular}{
    p{0.3cm}p{\widthof{Shin \etal~\cite{Shin2017} (\emph{A})}}
    p{0.3cm}p{\widthof{\emph{B}}}
    p{0.3cm}p{\widthof{\emph{C}}}}
    \ref*{pgfplots:shin_ablation} & Shin \etal~\cite{Shin2017} (\emph{A}) & 
    \ref*{pgfplots:b_ablation} & \emph{B} & 
    \ref*{pgfplots:c_ablation} & \emph{C} \\ 
    \ref*{pgfplots:d_ablation} & \emph{D} &
    \ref*{pgfplots:g_ablation} & \multicolumn{3}{l}{\textbf{Our diff. JPEG} (\emph{E})} \\
    \end{tabular}};
    \spy on (1.3, 1.1) in node [above, dash pattern=, line width=0.2mm, black, fill=white] at (2.3, 0.2);
    \spy on (5.1, 0.6) in node [above, dash pattern=, line width=0.2mm, black, fill=white] at (5.1, 1.0);
\end{tikzpicture}

%% file: artwork/forward_performance_ste_small.tex
\begin{tikzpicture}[every node/.style={font=\footnotesize}]
    \node[anchor=center] at (0.42\columnwidth, -0.5) {JPEG quality};
	\begin{groupplot}[
        group style={
            group name=forwardperste,
            group size=2 by 2,
            ylabels at=edge left,
            horizontal sep=29pt,
            vertical sep=8pt,
        },
        enlargelimits=false,
        legend style={nodes={scale=0.5}},
        height=3.45cm,
        width=0.54\columnwidth,
        grid=both,
        xticklabel shift={-1pt},
        yticklabel shift={-1pt},
        xtick pos=bottom,
        ytick pos=left,
        grid style={line width=.1pt, draw=white, dash pattern=on 1pt off 1pt},
        major grid style={line width=.15pt,draw=gray!40, dash pattern=on 1pt off 1pt},
        minor tick num=1,
        xmin=1,
        xmax=99,
        ylabel shift=-4.5pt,
        xtick={
            1, 20, 40, 60, 80, 99
        },
        xticklabels={
            1, 20, 40, 60, 80, 99
        },
        ticklabel style = {font=\footnotesize},
        ]
        % SSIM curve
        \nextgroupplot[
        ylabel=SSIM $\uparrow$,
        ytick={0.98, 0.99, 1.0},
        yticklabels={0.98, 0.99, 1.0},
        ymin=0.98,   
        ymax=1,
        ylabel style={at={(-0.20,0.5)}},
        ]
        
        \addplot[color=tud6b, mark=*, mark size=0.3pt] table[x=cs, y=ssim] {ours_forward.dat};
        \addplot[name path=upper,draw=none] table[x=cs, y expr=\thisrow{ssim}+1*\thisrow{ssimstd}] {ours_forward.dat};
        \addplot[name path=lower,draw=none] table[x=cs, y expr=\thisrow{ssim}-1*\thisrow{ssimstd}] {ours_forward.dat};
        \addplot[fill=tud6b!50, fill opacity=0.35] fill between[of=upper and lower];

        \addplot[color=tud11b, mark=*, mark size=0.3pt] table[x=cs, y=ssim] {ours_ste_forward.dat};
        \addplot[name path=upper,draw=none] table[x=cs, y expr=\thisrow{ssim}+1*\thisrow{ssimstd}] {ours_ste_forward.dat};
        \addplot[name path=lower,draw=none] table[x=cs, y expr=\thisrow{ssim}-1*\thisrow{ssimstd}] {ours_ste_forward.dat};
        \addplot[fill=tud11b!50, fill opacity=0.35] fill between[of=upper and lower];
        
        % PSNR curve
        \nextgroupplot[
        ylabel=PSNR (dB) $\uparrow$,
        ytick={36, 40, 44, 48, 52},
        yticklabels={36, 40, 44, 48, 52},
        ymin=34,   
        ymax=53,
        ylabel style={at={(-0.13,0.5)}},
        ]
        
        \addplot[color=tud6b, mark=*, mark size=0.3pt] table[x=cs, y=psnr] {ours_forward.dat};\label{pgfplots:ours_forward_ste};
        \addplot[name path=upper,draw=none] table[x=cs, y expr=\thisrow{psnr}+1*\thisrow{psnrstd}] {ours_forward.dat};
        \addplot[name path=lower,draw=none] table[x=cs, y expr=\thisrow{psnr}-1*\thisrow{psnrstd}] {ours_forward.dat};
        \addplot[fill=tud6b!50, fill opacity=0.35] fill between[of=upper and lower];

        \addplot[color=tud11b, mark=*, mark size=0.3pt] table[x=cs, y=psnr] {ours_ste_forward.dat};\label{pgfplots:ours_ste_forward_ste};
        \addplot[name path=upper,draw=none] table[x=cs, y expr=\thisrow{psnr}+1*\thisrow{psnrstd}] {ours_ste_forward.dat};
        \addplot[name path=lower,draw=none] table[x=cs, y expr=\thisrow{psnr}-1*\thisrow{psnrstd}] {ours_ste_forward.dat};
        \addplot[fill=tud11b!50, fill opacity=0.35] fill between[of=upper and lower];
        
	\end{groupplot}

    \node[draw,fill=white, inner sep=0.5pt, anchor=center] at (0.42\columnwidth, 2.075) {\tiny
    \begin{tabular}{
    p{0.3cm}p{\widthof{\textbf{Our diff. JPEG} (\emph{E})}}
    p{0.3cm}p{\widthof{\textbf{Our diff. STE JPEG} (\emph{F})}}}
    \ref*{pgfplots:ours_forward_ste} & \textbf{Our diff. JPEG} (\emph{E}) & 
    \ref*{pgfplots:ours_ste_forward_ste} & \textbf{Our diff. STE JPEG} (\emph{F}) \\
    \end{tabular}};
\end{tikzpicture}

%% file: table/fgsm.tex
\renewcommand\arraystretch{0.98}
\begin{tabularx}{\columnwidth}{@{\extracolsep{\fill}}l@{\hskip 0.25em}S[table-format=2.2]S[table-format=2.2]S[table-format=2.2]S[table-format=2.2]S[table-format=2.2]S[table-format=2.2]@{}}
	\toprule
	& \multicolumn{3}{c}{Top-1 acc $\downarrow$} & \multicolumn{3}{c}{Top-5 acc $\downarrow$}  \\
    \cmidrule{2-4}\cmidrule{5-7}
	{Approach\hspace{0.8cm}$\q$ range $\to$\hspace{-0.15cm}} & {1-99} & {1-10} & {11-99} & {1-99} & {1-10} & {11-99} \\
	\midrule
	No attack & 66.83 & 33.36 & 71.01 & 85.91 & 53.10 & 90.01 \\
	\midrule
	\multicolumn{7}{@{}l}{\scriptsize \textit{FSGM with} $\epsilon=3$} \\
    \midrule
    Xing \etal~\cite{Xing2021} & 53.92 & 26.00 & 57.42 & 77.90 & 45.65 & 81.93 \\
    Xie \etal~\cite{Xie2022} & 41.48 & 20.07 & 44.15 & 66.53 & 38.52 & 70.03 \\
	Shin \etal~\cite{Shin2017} & \textit{36.62} & 16.40 & \textit{39.15} & \textit{60.19} & \textit{32.38} & \textit{63.67} \\
    \textbf{Our diff. JPEG} & \textbf{36.51} & \textbf{15.80} & \textbf{39.10} & \textbf{59.92} & \textbf{30.89} & \textbf{63.54} \\
    \textbf{Our diff. STE JPEG} & 36.95 & \textit{16.35} & 39.53 & 60.79 & 32.55 & 64.32 \\
    \midrule
    \multicolumn{7}{@{}l}{\scriptsize \textit{FSGM with} $\epsilon=9$} \\
    \midrule
    Xing \etal~\cite{Xing2021} & 51.25 & 27.64 & 54.20 & 75.24 & 47.41 & 78.72 \\
    Xie \etal~\cite{Xie2022} & 38.29 & 18.15 & 40.81 & 61.89 & 35.38 & 65.21 \\
	Shin \etal~\cite{Shin2017} & 35.01 & 15.04 & \textit{37.51} & 57.40 & \textit{29.70} & 60.86 \\
    \textbf{Our diff. JPEG} & \textit{35.00} & \textbf{14.65} & 37.55 & \textit{57.28} & \textbf{28.97} & \textit{60.82} \\
    \textbf{Our diff. STE JPEG} & \textbf{34.77} & \textit{14.88} & \textbf{37.25} & \textbf{57.24} & 29.92 & \textbf{60.65} \\
	\bottomrule
\end{tabularx}

%% file: table/ifgsm.tex
\renewcommand\arraystretch{0.98}
\begin{tabular*}{\columnwidth}{@{\extracolsep{\fill}}l@{\hskip 0.25em}S[table-format=2.2]S[table-format=2.2]S[table-format=2.2]S[table-format=2.2]S[table-format=2.2]S[table-format=2.2]@{}}
	\toprule
	& \multicolumn{3}{c}{Top-1 acc $\downarrow$} & \multicolumn{3}{c}{Top-5 acc $\downarrow$}  \\
    \cmidrule{2-4}\cmidrule{5-7}
	{Approach\hspace{0.8cm}$\q$ range $\to$\hspace{-0.15cm}} & {1-99} & {1-10} & {11-99} & {1-99} & {1-10} & {11-99} \\
	\midrule
	\multicolumn{7}{@{}l}{\scriptsize \textit{IFSGM with} $\epsilon=3$} \\
    \midrule
    Xing \etal~\cite{Xing2021} & 43.44 & 24.42 & 45.82 & 72.52 & 45.55 & 75.90 \\
    Xie \etal~\cite{Xie2022} & 25.30 & 14.72 & 26.63 & 46.55 & 31.47 & 48.43 \\
	Shin \etal~\cite{Shin2017} & 15.11 & 8.98 & 15.88 & 27.21 & 19.99 & \textit{28.11} \\
    \textbf{Our diff. JPEG} & \textbf{14.39} & \textbf{\hphantom{1}7.97} & \textbf{15.19} & \textbf{25.79} & \textbf{17.53} & \textbf{26.83} \\
    \textbf{Our diff. STE JPEG} & \textit{15.00} & \textit{\hphantom{1}8.35} & \textit{15.83} & \textit{27.07} & \textit{18.73} & 28.12 \\
    \midrule
    \multicolumn{7}{@{}l}{\scriptsize \textit{IFSGM with} $\epsilon=9$} \\
    \midrule
    Xing \etal~\cite{Xing2021} & 39.59 & 24.99 & 41.41 & 67.73 & 45.41 & 70.52 \\
    Xie \etal~\cite{Xie2022} & 15.03 & 8.70 & 15.82 & 27.34 & 19.21 & 28.35 \\
	Shin \etal~\cite{Shin2017} & \textit{\hphantom{1}6.89} & 4.99 & \textit{\hphantom{1}7.12} & \textit{12.64} & 10.47 & \textit{12.91} \\
    \textbf{Our diff. JPEG} & \textbf{\hphantom{1}6.54} & \textbf{\hphantom{1}4.09} & \textbf{\hphantom{1}6.85} & \textbf{11.96} & \textbf{\hphantom{1}8.32} & \textbf{12.41} \\
    \textbf{Our diff. STE JPEG} & 7.14 & \textit{\hphantom{1}4.14} & 7.51 & 13.11 & \textit{\hphantom{1}8.89} & 13.64 \\
	\bottomrule
\end{tabular*}

%% file: artwork/ifgsm_small.tex
\begin{tikzpicture}[every node/.style={font=\footnotesize}, spy using outlines={black, line width=0.80mm, dashed, dash pattern=on 1.5pt off 1.5pt, magnification=2.66, size=0.75cm, connect spies, fill=white}]
    \node[anchor=center] at (0.42\columnwidth, -0.5) {JPEG quality};
	\begin{groupplot}[
        group style={
            group name=ifsgm,
            group size=2 by 1,
            ylabels at=edge left,
            horizontal sep=29pt,
            vertical sep=14pt,
        },
        legend style={nodes={scale=0.5}},
        height=3.45cm,
        width=0.54\columnwidth,
        grid=both,
        xticklabel shift={-1pt},
        yticklabel shift={-1pt},
        xtick pos=bottom,
        ytick pos=left,
        grid style={line width=.1pt, draw=white, dash pattern=on 1pt off 1pt},
        major grid style={line width=.15pt,draw=gray!40, dash pattern=on 1pt off 1pt},
        minor tick num=1,
        xmin=1,
        xmax=99,
        ylabel shift=2.0pt,
        xtick={
            1, 20, 40, 60, 80, 99
        },
        xticklabels={
            1, 20, 40, 60, 80, 99
        },
        ticklabel style = {font=\footnotesize},
        ]
        % Full curve
        \nextgroupplot[
        ylabel=Accucacy (\%) $\downarrow$,
        ytick={0, 5, 10, 15, 20},
        yticklabels={0, 5, 10, 15, 20},
        ymin=0,   
        ymax=19,
        ylabel style={at={(-0.12,0.5)}},
        ]
        \addplot[color=tud3b, mark=*, mark size=0.3pt] table[x=cs, y=top1acc] {3_ifgsm_shin_resnet50.dat};\label{pgfplots:shin_ifsgm};
        \addplot[color=tud11b, mark=*, mark size=0.3pt] table[x=cs, y=top1acc] {3_ifgsm_ours_ste_resnet50.dat};\label{pgfplots:ours_ste_ifsgm};
        \addplot[color=tud6b, mark=*, mark size=0.3pt] table[x=cs, y=top1acc] {3_ifgsm_ours_resnet50.dat};\label{pgfplots:ours_ifsgm};
        
        % Low CS curve
        \nextgroupplot[
        ylabel=Accucacy (\%) $\downarrow$,
        ytick={0, 2, 4, 6, 8},
        yticklabels={0, 2, 4, 6, 8},
        ymin=0,   
        ymax=9,
        xmin=1,   
        xmax=5,
        xtick={
            1, 2, 3, 4, 5
        },
        xticklabels={
            1, 2, 3, 4, 5
        },
        ylabel style={at={(-0.12,0.5)}},
        ]
        \addplot[color=tud3b, mark=*, mark size=0.5pt] table[x=cs, y=top1acc] {3_ifgsm_shin_resnet50.dat};
        \addplot[color=tud11b, mark=*, mark size=0.5pt] table[x=cs, y=top1acc] {3_ifgsm_ours_ste_resnet50.dat};
        \addplot[color=tud6b, mark=*, mark size=0.5pt] table[x=cs, y=top1acc] {3_ifgsm_ours_resnet50.dat};
	\end{groupplot}
    
    \node[draw,fill=white, inner sep=0.5pt, anchor=center] at (0.41\columnwidth, 2.075) {\tiny
    \begin{tabular}{
    p{0.3cm}p{\widthof{Shin \etal \cite{Shin2017}}}
    p{0.3cm}p{\widthof{Xing \etal \cite{Xing2021}}}
    p{0.3cm}p{\widthof{Xing \etal \cite{Xing2021}}}}
    \ref*{pgfplots:ours_ifsgm} & \textbf{Ours} &  \ref*{pgfplots:ours_ste_ifsgm} & \textbf{Ours STE} & \ref*{pgfplots:shin_ifsgm} &
    Shin \etal \cite{Shin2017}
    \end{tabular}};
    \spy on (2.0, 1.59) in node [above, dash pattern=, line width=0.2mm, black, fill=white] at (2.0, 0.2);
    % \spy on (6.55, 1.69) in node [above, dash pattern=, line width=0.2mm, black, fill=white] at (6.25, 0.2);
\end{tikzpicture}

%% file: table/vanishing_gradient.tex
\renewcommand\arraystretch{0.98}
\begin{tabular*}{\columnwidth}{@{\extracolsep{\fill}}l@{\hskip 0.1em}S[table-format=1.3]S[table-format=1.3]S[table-format=1.3]S[table-format=1.3]S[table-format=1.3]S[table-format=1.3]@{}}
	\toprule
	& \multicolumn{3}{c}{$\mathbb{E}_{\mathrm{Q},\mathcal{D}}[\|\Delta_{\q}\mathcal{L}_{1}\|]$} & \multicolumn{3}{c}{$\!\!\!\mathbb{E}_{\mathrm{Q},\mathcal{D}}[\|\Delta_{\qt_{Y, C}}\mathcal{L}_{1}\|]\!\!\!$}  \\
    \cmidrule{2-4}\cmidrule{5-7}
	{Approach\hspace{0.4cm}$\q$ range $\to$\hspace{-0.15cm}} & {1-99} & {1-10} & {11-99} & {1-99} & {1-10} & {11-99} \\
	\midrule
    Xing \etal~\cite{Xing2021} & 1.254 & 7.271 & 0.502 & {--} & {--} & {--} \\
    Xie \etal~\cite{Xie2022} & 0.955 & 4.617 & 0.492 & 0.913 & 1.041 & 0.896 \\
	Shin \etal~\cite{Shin2017} & 0.587 & 4.172 & 0.139 & 0.213 & 0.474 & 0.181 \\
    \textbf{Our diff. JPEG} & \textit{0.022} & \textit{0.068} & \textit{0.017} & \textbf{0.043} & \textbf{0.030} & \textbf{0.044} \\
    \textbf{Our diff. STE JPEG} & \textbf{0.014} & \textbf{0.042} & \textbf{0.010} & \textit{0.060} & \textit{0.162} & \textit{0.048} \\
	\bottomrule
\end{tabular*}

%% file: table/round_function_forward_function.tex
\renewcommand\arraystretch{0.95}
\begin{tabular*}{\columnwidth}{@{\extracolsep{\fill}}l@{\hskip 0.1em}S[table-format=1.3]S[table-format=1.3]S[table-format=1.3]S[table-format=2.2]S[table-format=2.2]S[table-format=2.2]@{}}
	\toprule
	& \multicolumn{3}{c}{SSIM $\uparrow$} & \multicolumn{3}{c}{PSNR $\uparrow$}  \\
    \cmidrule{2-4}\cmidrule{5-7}
	{Function\hspace{1.85cm}$\q$ range $\to$\hspace{-0.4cm}} & {1-99} & {1-10} & {11-99} & {1-99} & {1-10} & {11-99} \\
	\midrule
    Fourier \textcolor{tud0d}{{\fontsize{4}{5}\selectfont$x-\sum^{10}_{k=1}\!\!\!\frac{(-1)^{k+1}}{k\pi}\!\sin(2\pi kx)\!$}} & 0.984 & 0.956 & 0.987 & 41.28 & 35.66 & 41.98 \\
    Linear \textcolor{tud0d}{{\fontsize{4}{5}\selectfont$\lfloor x\rceil+0.1\left(x - \lfloor x\rceil\right)$}} & \textit{0.990} & \textbf{0.987} & 0.991 & \textit{42.37} & \textbf{38.70} & 42.83 \\
    Polynomial \textcolor{tud0d}{{\fontsize{4}{5}\selectfont$\lfloor x\rceil+(x - \lfloor x\rceil)^{3}$}} & \textbf{0.991} & \textit{0.987} & \textbf{0.992} & \textbf{42.60} & \textit{38.28} & \textbf{43.14} \\
    Sigmoid \textcolor{tud0d}{{\fontsize{4}{5}\selectfont$\sigma(60(x-\frac{1}{2}+\lfloor x\rfloor)) + \lfloor x\rfloor$}} & 0.990 & 0.980 & \textit{0.992} & 42.29 & 36.88 & \textit{42.96} \\
    Tanh \textcolor{tud0d}{{\fontsize{4}{5}\selectfont$\frac{1}{2}\tanh\!\left(5\left(x - \frac{1}{2} - \lfloor x\rceil\right)\right)+\frac{1}{2}$}} & 0.972 & 0.918 & 0.978 & 38.51 & 31.15 & 39.42 \\
	\bottomrule
\end{tabular*}
%  $\sigma\left(60(x-0.5+\lfloor x\rfloor)\right) + \lfloor x\rfloor$
% $\lfloor x\rceil+0.1\left(x - \lfloor x\rceil\right)$

%% file: table/round_function_ifgsm.tex
\renewcommand\arraystretch{0.98}
\begin{tabular*}{\columnwidth}{@{\extracolsep{\fill}}l@{\hskip 0.25em}S[table-format=2.2]S[table-format=2.2]S[table-format=2.2]S[table-format=2.2]S[table-format=2.2]S[table-format=2.2]@{}}
	\toprule
	& \multicolumn{3}{c}{Top-1 acc $\downarrow$} & \multicolumn{3}{c}{Top-5 acc $\downarrow$}  \\
    \cmidrule{2-4}\cmidrule{5-7}
	{Function\hspace{0.8cm}$\q$ range $\to$\hspace{-0.15cm}} & {1-99} & {1-10} & {11-99} & {1-99} & {1-10} & {11-99} \\
	\midrule
    Fourier & 39.53 & 20.16 & 41.95 & 68.98 & 40.81 & 72.50 \\
    Linear & 25.69 & 22.41 & 26.10 & 46.52 & 42.84 & 46.98 \\
    Polynomial & \textbf{14.39} & \textit{\hphantom{1}7.97} & \textbf{15.19} & \textbf{25.79} & \textit{17.53} & \textbf{26.83} \\
    Sigmoid & \textit{20.28} & \textbf{\hphantom{1}6.34} & \textit{22.02} & \textit{36.79} & \textbf{14.44} & \textit{39.59} \\
    Tanh & 22.52 & 15.20 & 23.43 & 41.80 & 32.79 & 42.92 \\
	\bottomrule
\end{tabular*}

%% file: table/ste_backward_function.tex
\renewcommand\arraystretch{0.98}
\begin{tabular*}{\columnwidth}{@{\extracolsep{\fill}}l@{\hskip 0.25em}S[table-format=2.2]S[table-format=2.2]S[table-format=2.2]S[table-format=2.2]S[table-format=2.2]S[table-format=2.2]@{}}
	\toprule
	& \multicolumn{3}{c}{Top-1 acc $\downarrow$} & \multicolumn{3}{c}{Top-5 acc $\downarrow$}  \\
    \cmidrule{2-4}\cmidrule{5-7}
	{Backw. approach\hspace{0.35cm}$\q$ range $\to$\hspace{-0.15cm}} & {1-99} & {1-10} & {11-99} & {1-99} & {1-10} & {11-99} \\
	\midrule
    Constant grad. (standard STE) & 25.30 & 21.62 & 25.76 & 45.38 & 41.37 & 45.88 \\
    \textbf{Surrogate (ours)} & \textbf{\hphantom{1}7.14} & \textbf{\hphantom{1}4.14} & \textbf{\hphantom{1}7.51} & \textbf{13.11} & \textbf{\hphantom{1}8.89} & \textbf{13.64}  \\
	\bottomrule
\end{tabular*}

%% file: content/conclusion.tex
\section{Conclusion}
\label{sec:conclusion}

We reviewed existing differentiable JPEG approaches and proposed a novel differentiable JPEG approach modeling missing properties of standard (non-diff.) JPEG. Our approach is the first to accurately resemble standard JPEG over the entire JPEG quality range, outperforming all existing approaches. Notably, gradients derived from our approach yield superior adversarial examples, demonstrating the ``usefulness'' of the obtained gradients for gradient-based optimization. With strong forward and backward performance, our approach provides a solid foundation for future work considering JPEG as part of deep vision pipelines.